\documentclass[runningheads]{llncs}

\usepackage{eccv}

\usepackage{eccvabbrv}

\usepackage{graphicx}
\usepackage{booktabs}
\usepackage{wrapfig}

\usepackage[accsupp]{axessibility}  

\usepackage{hyperref}

\usepackage{orcidlink}

\hypersetup{
    colorlinks,
    linkcolor={red!80!black},
    citecolor={green!80!black},
    urlcolor={blue!80!black}
}

\usepackage{subcaption}
\usepackage[table]{xcolor}         
\definecolor{tablegray}{gray}{0.92}

\usepackage{stackengine}
\usepackage{xcolor}

\newcommand{\inatlabel}[1]{%
  \bgroup\setlength{\fboxsep}{1.5pt}%
  \colorbox{black!70}{
  \parbox{1.85cm}{
  \raggedleft
  \tiny\color{white} #1}
  }\egroup
}

\newcommand{\method}{\textsc{WildProp}\xspace}
\newcommand{\avonet}{AVONET\xspace}
\newcommand{\shore}{ShoreBirds\xspace}
\newcommand{\frogs}{Frogs\xspace}

\newcommand{\customparagraph}[1]{\medskip\noindent\textbf{#1}\quad}
\newcommand{\customshortparagraph}[1]{\smallskip\noindent\textbf{#1}\quad}

\usepackage{multirow}
\usepackage{algorithm}
\usepackage{algpseudocode}
\usepackage{amsmath}

\begin{document}

\title{\method: Visual Estimation of Wildlife Body Proportions at Scale}

\author{Mustafa Chasmai,~Aaron Sun,~Subhransu Maji}
\authorrunning{Chasmai et al.}

\institute{\email{\{mchasmai,~aaronsun,~smaji\}@umass.edu}\\Manning College of Information and Computer Sciences\\University of Massachusetts, Amherst}

\maketitle

\begin{figure}
    \centering
    \vspace{-2mm}
    \includegraphics[width=0.94\linewidth]{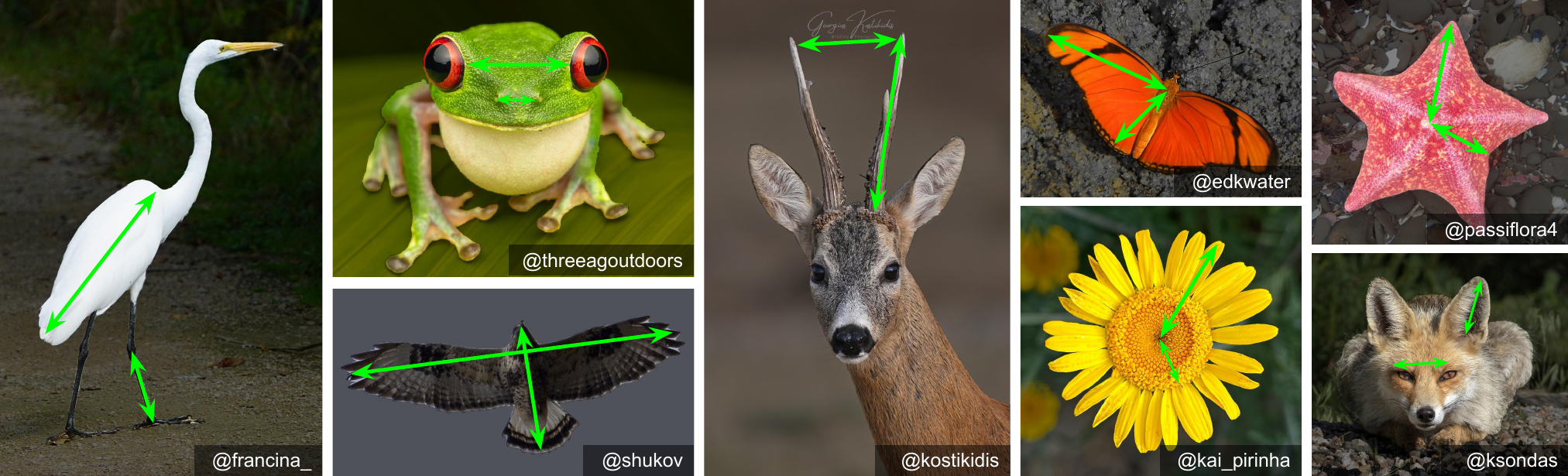}
    \vspace{-2mm}
    \caption{\textbf{Estimating wildlife body proportions.} Given a single annotated reference image and a set of specified body parts, \method estimates the population-level distribution of their length ratios from a large, uncurated image collection. We show exemplar species and body parts used in our experiments. All images are sourced from iNaturalist~\cite{inat}; observer IDs are listed below each image.}
    \label{fig:main_teaser}
    \vspace{-8mm}
\end{figure}

\begin{abstract}
Population-level morphometric measurements are central to ecological and evolutionary studies, but traditionally require controlled imaging or physical specimen handling, limiting their scalability. We present \method, a framework for estimating wildlife body-proportion distributions directly from large-scale, unconstrained image repositories. We cast morphometric estimation as a retrieval-driven correspondence problem: given a single user-annotated canonical image, \method performs pose-aware retrieval using foundation model features, transfers part endpoints via dense patch-level matching, filters predictions using geometric consistency, and aggregates measurements across retrieved images to estimate length-ratio distributions. Unlike supervised keypoint pipelines, our approach adapts to arbitrary species and user-defined body parts without per-species training. Evaluations on three large morphometric datasets spanning birds and amphibians show median relative errors of 10--20\%. We further demonstrate broad applicability through case studies measuring proportions across diverse taxa, including birds, frogs, insects, and flowers. Ablations show that pose-aware retrieval is critical for stable estimation, while robust aggregation mitigates keypoint and pose noise. Our results suggest that carefully filtered 2D correspondences over web-scale imagery can provide scalable morphometric proxies for comparative analyses across taxa, geography, and seasonality.
\end{abstract}

\section{Introduction}
\label{sec:intro}
Quantitative morphometric measurements---such as ratios of limb, wing, or head dimensions---are central to ecology and evolutionary biology (see \cref{fig:main_teaser}). Large-scale efforts such as AVONET~\cite{tobias2022avonet} have demonstrated how curated trait datasets can inform macroevolution~\cite{fitzjohn2010quantitative}, community assembly~\cite{laughlin2014intrinsic}, ecosystem modeling~\cite{harfoot2014emergent}, and population health assessment~\cite{hodgson2020rapid, kershaw2017evaluating, morfeld2014development}. Morphological variation within species across geography, and convergence across species within shared environments, are often diagnosed through such proportional measurements. These patterns have long served as the empirical basis for fundamental biological frameworks such as the theory of evolution~\cite{darwin1859}.

Collecting morphometric measurements at scale, however, remains difficult. Traditional pipelines rely on physical specimen handling or controlled photography with standardized poses and rulers (see \cref{fig:teaser}). These approaches can be precise, but are labor-intensive, disruptive to wildlife and ecosystems, and limited in coverage; many datasets contain measurements for only a small number of individuals per species. In contrast, public repositories such as iNaturalist~\cite{inat} contain millions of wildlife images spanning taxa, geography, and time. These collections offer an appealing opportunity for scalable morphometric analysis, but geometric measurement from arbitrary photographs is ill-posed due to viewpoint variation, articulation, occlusion, scale ambiguity, and perspective distortion.

Our key observation is that although most images in such repositories are unsuitable for direct measurement, web-scale collections often contain a substantial subset of images captured in near-canonical poses where relative geometric relationships are approximately preserved. This suggests a different formulation: rather than reconstructing geometry from arbitrary views, we can selectively retrieve measurement-friendly instances and aggregate their statistics.

We introduce \method, a scalable, training-free framework for estimating population-level body-proportion distributions from large, unconstrained image collections (see \cref{fig:method}). Given a single reference image in a canonical pose with user-defined part endpoints, \method: (1) retrieves pose-aligned images from the desired population using foundation-model representations; (2) transfers part endpoints via dense patch-level matching with geometric consistency filtering; and (3) aggregates measurements through iterative refinement to estimate stable ratio distributions. This formulation is user-friendly and flexible: new species and traits require only a single annotated query image, without per-species training or task-specific keypoint labels. Recent advances in self-supervised visual representations, such as DINO~\cite{simeoni2025dinov3}, and foundation segmentation models, such as SAM~\cite{sam}, make such training-free correspondence robust across diverse taxa.

We evaluate \method on three large morphometric datasets spanning birds and amphibians, mimicking the original measurement protocols and comparing estimated population ratios to physical ground truth. Given only a single annotated image, \method achieves 10–20\% median relative error on various part-length ratio estimation tasks using large-scale image collections. Ablations demonstrate that viewpoint-aware retrieval, geometric verification, and large retrieval sets are critical for estimation accuracy.
We further present case studies illustrating broad applicability across taxa and subgroup analyses across geography and seasonality. 
We conclude with a discussion of limitations and directions for future work. Code available at \href{https://github.com/cvl-umass/wildprop}{github:cvl-umass/wildprop}.

\section{Related Work}
\label{sec:related}

\begin{figure}[t]
    \centering
    \includegraphics[width=0.98\linewidth]{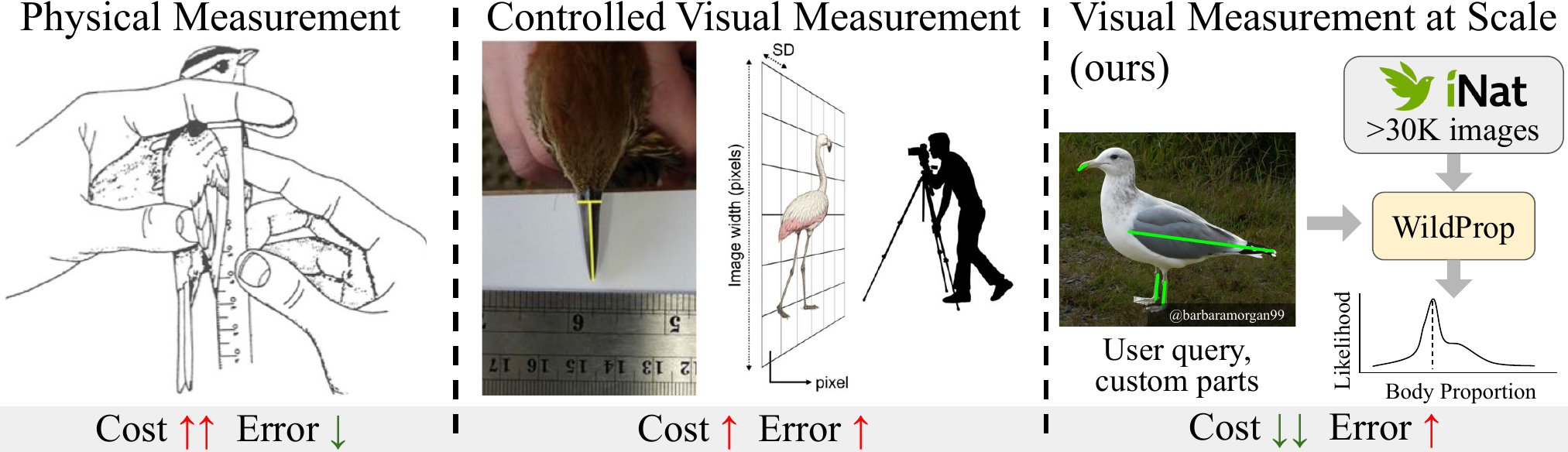}
    \caption{\textbf{Morphometry paradigms\protect\footnotemark.} The choice between the measurement method is driven by a tradeoff between measurement costs and estimation errors.}
    \label{fig:teaser}
    \vspace{-4mm}
\end{figure}

\footnotetext{The first 3 images are borrowed from \avonet~\cite{tobias2022avonet},  \cite{williams2020photography} and \cite{delfino2026photogrammetry}, respectively.}

\customshortparagraph{Morphometrics.}
Measuring morphological traits has a long history in ecology and evolutionary biology, where the shape, size, and structure of body parts are used to study species and their interactions within ecosystems~\cite{bartomeus2016common, mayfield2010opposing, schleuning2015predicting}. A classic example comes from Darwin’s theory of evolution~\cite{darwin1859}: variation in finch beak morphology in the Gal\'apagos Islands provided key evidence for natural selection. Morphometric traits can reveal evolutionary adaptations~\cite{conklin2019wry}, help identify gender in some species~\cite{cooke2020photography, pazvant2022sex}, and distinguish visually similar species~\cite{lee2025identification}. Large-scale studies have measured morphological traits across diverse taxa, including plants~\cite{kattge2020try, funk2017revisiting}, mammals~\cite{jones2009pantheria, wilman2014eltontraits, myhrvold2015amniote, fish, mussels}, amphibians~\cite{frogs}, insects~\cite{gao2024morphological}, and birds~\cite{ricklefs1980morphological, tobias2022avonet, shore}. Global coverage of such measurements could enable the development of general ecosystem models~\cite{harfoot2014emergent}, but the scale and diversity of species make this task extremely challenging. The lack of standardized measurement frameworks across taxonomic groups further exacerbates the problem. Beyond broad coverage, stratification is another key aspect of morphometric studies. Repeated measurements over time can reveal ecological change~\cite{youngflesh2022abiotic,gomez2025evidence,phillips2024analysis, geladi2019100}, while analyses across demographic or geographic subgroups can provide insights into species variation~\cite{russell2023comparisons} and population health~\cite{hodgson2020rapid, kershaw2017evaluating, morfeld2014development}. However, such studies are typically limited to specific niches and are difficult to scale globally. 
Our approach, \method, scales to global taxonomic diversity, readily adapts to user-specified body proportions, and enables flexible stratification by appropriately restricting the retrieval set.

\customparagraph{Vision for morphometrics.}
Prior efforts to measure body size and proportions tend to rely on physical, in-situ studies or controlled photography (\cref{fig:teaser}), followed by digital measurement in an effort to reduce the required domain expertise and maintain persistent records for cross-verification. Compared to traditional physical measurement protocols, these approaches report significantly reduced specimen handling time~\cite{williams2020photography}, which can enable faster data collection and improved scalability. 
This is a practical and scalable setup for livestock~\cite{ma2024computer, peng2024automated} and museum specimens~\cite{he2023using, weeks2023deep, he2025phenolearn}, but is quite cost-intensive for wildlife in natural ecosystems~\cite{williams2020photography, kristiansen2025husmorph, panda2024digital, delfino2026photogrammetry, cooke2020photography}. Efforts towards non-invasive alternatives, such as camera traps~\cite{fergus2024towards} and aerial imagery~\cite{bagchi2025automated}, have generally been limited to large mammals. 
In contrast, our approach is highly scalable, completely non-invasive and broadly applicable to diverse taxonomic groups.

\customparagraph{Animal pose estimation.} Animal pose estimation has received significant attention in recent years, building on advances in computer vision~\cite{graving2019deepposekit, xiong2025diffpose, rao2025probabilistic, cao2019cross, lu2022few}. However, these methods typically assume fixed keypoint definitions and require extensive annotations. A notable example is DeepLabCut~\cite{mathis2018deeplabcut}, which has been widely used for detection and tracking of animals but still relies on manual annotation and training on the order of 100–200 images per category. We note that even perfect 2D localization does not resolve measurement ambiguity due to scale and foreshortening effects arising from projection. While monocular depth estimation has improved substantially~\cite{yang2024depth, piccinelli2025unidepthv2, ke2024repurposing}, existing models are optimized for large rigid objects and background geometry, and are unreliable for fine-grained articulated structures typical in wildlife imagery. 
General image-to-3D reconstruction has also advanced rapidly in recent years, including foundation models (e.g., SAM3D~\cite{chen2026sam}, Trellis~\cite{xiang2025structured, xiang2026native}) and recent extensions to animal 3D reconstruction~\cite{hu2026sam}; however, these methods have not yet been widely studied for precise morphometric measurement.
Instead of explicitly solving the 3D measurement problem, we mitigate perspective effects by conditioning on pose through retrieval, effectively restricting viewpoint variability before estimation. This also reduces the burden on the keypoint matching step, allowing us to generalize from a single example using features from modern foundation models~\cite{simeoni2025dinov3,oquab2023dinov2}.

\customparagraph{CV for ecology.} Computer vision has been widely applied to ecological problems~\cite{tuia2026towards}, driven in part by large community-driven platforms such as iNaturalist~\cite{van2018inaturalist,cole2022label,inat}, eBird, and others. Recent work has also explored the use of large vision–language models~\cite{vendrow2024inquire} to enable ecologists to query arbitrary attributes beyond species identity. 
Automated vision approaches have already seen some success in large scale investigations of qualitative traits, such as species coloration and plumage~\cite{hantak2022computer, cooney2022latitudinal}. 
Our work on quantitative morphological measurements is complementary to these efforts. For example, attributes such as age, sex, color, or geographic location could be used to further stratify population-level estimates beyond species. We investigate several such examples in our case studies.

\section{Methodology}
\label{sec:methods}

\begin{figure}[t]
    \centering
    \includegraphics[width=\linewidth]{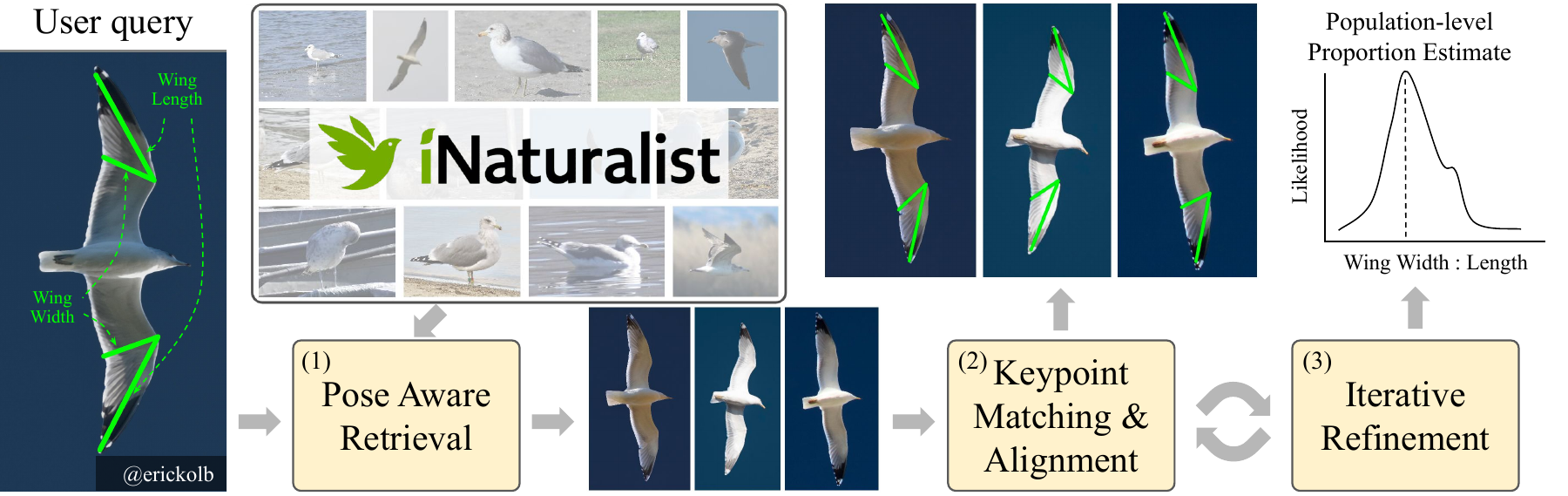}
    \caption{\textbf{Method overview.} The proposed approach consists of three key components: (1) pose-aware retrieval to identify images from a large collection (\eg, iNaturalist) where the desired measurements can be reliably estimated; (2) keypoint matching and geometric alignment to transfer part annotations; and (3) iterative refinement of keypoint appearance and correspondences to estimate population-level body proportions. In this example, the user provides a canonical image with annotated keypoints defining the ratio of wing width to length (left). The top three pose-aligned retrieved images and their estimated correspondences are shown in the middle panels, from which the distribution of wing width–to–length ratios is computed. Also, note that in this case, due to symmetry, each instance provides two measurements of width and length.}
    \label{fig:method}
    \vspace{-3mm}
\end{figure}

The overall approach is illustrated in Fig.~\ref{fig:method}. The user provides a single annotated image (query) with multiple specified body parts (at least two). We assume that these parts are approximately fronto-parallel in the image, so that foreshortening effects are negligible. The goal is to estimate the ratio of these body parts from a large collection of relevant images (e.g., images of the same species or further stratified subsets). Our approach consists of three steps: (1) retrieving images from the collection with object poses similar to the query image; (2) performing keypoint matching and alignment; and (3) iteratively refining appearance and alignment to estimate population-level distributions of part-length ratios.

\subsection{User Query}
The user provides a query image $I_q$, a set of keypoints $K=\{(x_1,y_1), \ldots, (x_k,y_k)\}$, and a set of parts $P=\{(i_1,j_1), \ldots, (i_n,j_n)\}$, where $1 \leq i_\ell,j_\ell \leq k$ and $n \geq 2$. Each part $(i_\ell,j_\ell)$ specifies the two keypoints defining a body segment, and $(x_i,y_i)$ denotes the image location of the $i$-th keypoint. The goal is to estimate ratios between the lengths of the specified parts over a target image collection.

\customparagraph{Discussion.} \method assumes that the annotated parts are approximately fronto-parallel in the chosen view. This assumption may not hold when the parts needed for a ratio are best observed from different viewpoints (\eg bill length from a side view and bill width from a frontal view). In such cases, the ratio can still be estimated by using a common anchor part visible across views (\eg wing or body length) and applying \method with multiple query images. For example, one can estimate each part relative to the shared anchor in its appropriate view, and then combine these estimates to recover the desired cross-view ratio.

\subsection{Pose-aware Retrieval}

We assume access to a large image collection representing the target population. In most experiments, this collection consists of high-resolution iNaturalist images~\cite{inat} of the same species, though we also perform analyses stratified by season and geographic region. The images are captured in the wild and therefore vary substantially in pose, viewpoint, background, scale, and image quality. We do not assume access to camera calibration information.

We first retrieve images whose poses are similar to that of the query image. Patch-level DINOv3 embeddings are effective for pose-aware retrieval, but can be sensitive to background features. To reduce background influence, we segment the target object using SAM2~\cite{sam} together with Grounding-DINO~\cite{liu2023grounding}, using simple text prompts such as ``bird'' or ``frog,'' and zero-out the background. Each segmented image is resized to $518\times518$, and DINOv3 patch embeddings are extracted and concatenated into a single image descriptor. We then use cosine similarity between descriptors as the retrieval criterion. We explore alternative retrieval variants in \cref{tab:abl:retrieval}. In our experiments, we retain the top 100 matches for each query. Although this is only a small subset of the full collection, it is often substantially larger than the number of individuals in traditional morphometric datasets. For example, \avonet~\cite{tobias2022avonet} reports fewer than nine individuals per species on average.

\customparagraph{Motivation.} The goal of this stage is to efficiently retrieve a small set of candidate images from a large collection, typically containing $10^4$--$10^5$ images. We therefore use global image-level descriptors for fast retrieval, and defer fine-grained keypoint alignment and geometric verification to the next stage. This coarse-to-fine design parallels classical SIFT-based instance retrieval pipelines in computer vision~\cite{sivic2003video, nister2006scalable}, where efficient global retrieval is followed by local feature matching and geometric verification.

\subsection{Keypoint Matching and Alignment}
For each retrieved image, we transfer the part annotations from the query image through patch-level keypoint matching. We construct background-suppressed images by halving the intensity of pixels identified as background by SAM2~\cite{sam} and extract their DINOv3~\cite{simeoni2025dinov3} patch embeddings. For each query keypoint $(x_i,y_i)$, we extract the 768-dimensional embedding of the nearest $16\times16$ patch in the query image $I_q$. We then find the patch in the retrieved image with the highest cosine similarity and use its center as the predicted corresponding location $(x'_i,y'_i)$. In our ablation studies, we find that higher-resolution inputs improve matching accuracy, but the gains tend to saturate beyond 518$\times$518.

Individual matches can be error-prone, especially for symmetric or visually similar parts such as wing tips, legs, or repeated body structures. We therefore apply a geometric consistency check. Given the predicted correspondences, we use RANSAC~\cite{ransac} to fit a rigid transformation between the query keypoints and their predicted locations in the retrieved image. Images with more than a specified number of outlier correspondences are excluded from estimation. We use a relatively permissive outlier threshold to account for natural variation in shape and pose across individuals.

\subsection{Iterative Refinement and Ratio Estimation}

Keypoint appearance features extracted from a single query image may not capture the visual diversity of the target population, leading to alignment errors. To address this, we introduce an iterative refinement step that adapts the keypoint descriptors to the retrieved image set. Given the correspondences estimated in the previous step, we update each keypoint descriptor by averaging the corresponding patch features from the top $K$ retrieved images together with the original query feature. The parameter $K$ controls the degree of adaptation to the target distribution. This refinement is particularly useful when transferring annotations across species or across visually diverse subpopulations.

The updated keypoint descriptors are then used to re-estimate correspondences in the retrieved images, and the process is repeated for a fixed number of iterations. After refinement, we compute the length of each specified part in every retained image and form the corresponding part-length ratios. These per-image ratios are then aggregated to estimate population-level statistics such as the mean, median, and variance.

\subsection{Implementation Details}
We use the ViT-B/16 DINOv3 backbone~\cite{simeoni2025dinov3} for both retrieval and keypoint matching, and the ViT-H variant of SAM2~\cite{sam} for masking. For iterative refinement, we perform three iterations. At each iteration, we update keypoint descriptors using the top 20 retrieved images together with the original query descriptors, aggregated by a weighted average with weights proportional to feature similarity. These hyperparameters were selected based on relative error on the AVONET California Gull validation setting~\cite{tobias2022avonet} and then applied unchanged to all other species and datasets.

For RANSAC, we use a fitting threshold of 5\% of the image diagonal and discard a retrieved image if more than 20\% of the predicted keypoints are classified as outliers. Additional hyperparameter details are provided in Appendix~A1. For the CLIP variants explored in \cref{sec:ablation}, we use the ViT-B/16 backbone. For Stable Diffusion~\cite{esser2024scaling}, we use the v3.5-medium model. All experiments were conducted on a single NVIDIA A100 GPU.

\section{Experiments and Results}
\label{sec:results}

\subsection{Datasets and Evaluation Metrics}

We validate this approach on three large morphometric datasets spanning birds and amphibians, as well as several case studies involving other taxa, and show that large image collections on platforms such as iNaturalist, combined with representations from existing foundation models, enable scalable, population-level estimation of body proportions. 

For each experiment, we perform retrieval on full-resolution images from iNaturalist~\cite{inat} that (1) belong to the search corpus, (2) were captured on or before 31 December 2025, and (3) are licensed for research use. We also manually search for suitable query images on iNaturalist and label keypoints for the various body parts using Labelbox~\cite{labelbox}. Example species and body parts used in our experiments are shown in~\cref{fig:main_teaser}.

For our quantitative experiments, part specifications vary across datasets, and we annotate keypoints according to the prescribed measurement protocols for each dataset. Consider \avonet~\cite{tobias2022avonet}, for example (see~\cref{tab:avonet}, right): we measure bill length by placing one keypoint at the tip of the bill and another at the anterior edge of the nostrils (nares). Some parts are not feasible to measure visually---for example, \avonet includes another bill measurement from the tip to the base of the skull, which is typically covered by feathers---and we therefore exclude such measurements from our experiments. Each dataset reports linear measurements for individual specimens across various species, and we use proportions derived from these measurements as references for our estimates.

We generally compute all part proportions relative to the longest part (\eg wing length for birds) for numerical stability. Factors such as age, gender, and health introduce variance in measurements across individuals of a species, and we aim to capture the overall distribution of each body proportion. In the absence of individual correspondence, we compare summary statistics. Concretely, we compute the median proportion estimated by \method ($\hat{\mu}$) 
and the median proportion measured in the dataset ($\mu$), and report the relative estimation error as $E = \left|\hat{\mu} - \mu\right|/\mu$. 

For our qualitative case studies, we explore diverse taxonomic groups and select species with distinct morphological patterns that can be qualitatively verified. For example, we analyze butterfly species with substantially different wing dimensions and evaluate whether the estimated proportions separate them. We use kernel density estimation~\cite{silverman1986density} to visualize the estimated distributions.

\subsection{Quantitative Results}

\cref{tab:avonet,tab:shore,tab:frogs} show quantitative comparisons with existing physical morphometric studies on  birds~\cite{tobias2022avonet}, shorebirds~\cite{shore}, and frogs~\cite{frogs}, respectively. 
Query images used for the different canonical poses are shown next to each table, with all measured parts visualized as green lines. 
We also visualize selected model predictions from these datasets in \cref{fig:pred_vis} and the Appendix~A5.

\begin{wrapfigure}{r}{0.49\textwidth}
  \centering
  \includegraphics[width=0.48\textwidth]{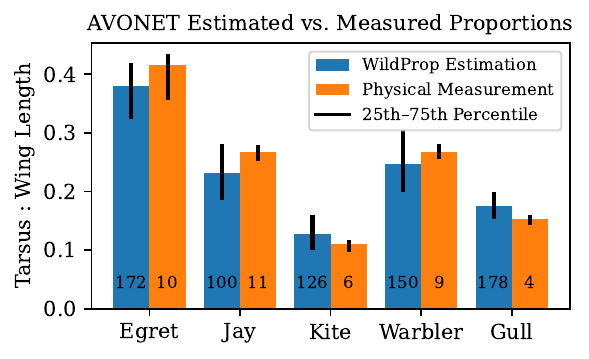} 
    \vspace{-2mm}
  \caption{Tarsus:Wing-length ratio medians and quartiles  across \avonet species. Sample sizes for both \method estimations and physical measurements are indicated at the base of each bar. }
  \label{fig:quantiles}
  \vspace{-6mm}
\end{wrapfigure}

\customparagraph{\avonet.} \cref{tab:avonet} shows results on \avonet across five species spanning several bird families. Following the measurement protocol outlined by the authors, we estimate four parts across two poses: bill length, tarsus (leg length), wing width, and wing length. On average, our measurements have about 11\% relative error compared to the ground truth median estimates.
\cref{fig:quantiles} shows the median and the 25--75 quantiles estimated using our approach and the ground truth across species. While the medians follow similar trends, our estimates have larger interquartile ranges. However, given the small sample sizes ($n\leq11)$ in these studies, it is unclear whether the physical measurements capture the full variation in the population.

\begin{table}[t]
    \centering
    \setlength{\tabcolsep}{3pt}
    \caption{\textbf{\avonet~\cite{tobias2022avonet}.} Relative error (\%) in median proportions. Lower is better. To the right are example query images and keypoints: side view for Cape May Warbler (bill, tarsus, and wing); and bottom view for California Gull (wing length and width).}
    \label{tab:avonet}
    \vspace{-2mm}
    \begin{minipage}{0.76\textwidth}
        \begin{tabular}{lcccc}
        \toprule
            \multirow{2}{*}{Species} & Bill L.  & Tarsus & Wing W. & \multirow{2}{*}{Mean} \\
            & / Wing L. & / Wing L. & / Wing L. \\
            \midrule
            Great Egret & 16.7 & 8.6 & 9.0 & 11.4\\
            Blue Jay & 4.0 & 12.9 & 8.8 & 8.6\\
            White-tailed Kite & 15.9 & 15.7 & 7.7 & 13.1\\
            Cape May Warbler & 14.1 & 8.1 & 15.2 & 12.5\\
            California Gull & 0.4 & 14.9 & 18.0 & 11.1\\
            \midrule
            Mean & 10.2 & 12.0 & 11.7 & 11.3\\
        \bottomrule
        \end{tabular}
    \end{minipage}
    \begin{minipage}{0.22\textwidth}
        
        \stackinset{r}{3pt}{b}{0pt}{\inatlabel{@animalmanpa}}{%
            \includegraphics[width=\linewidth]{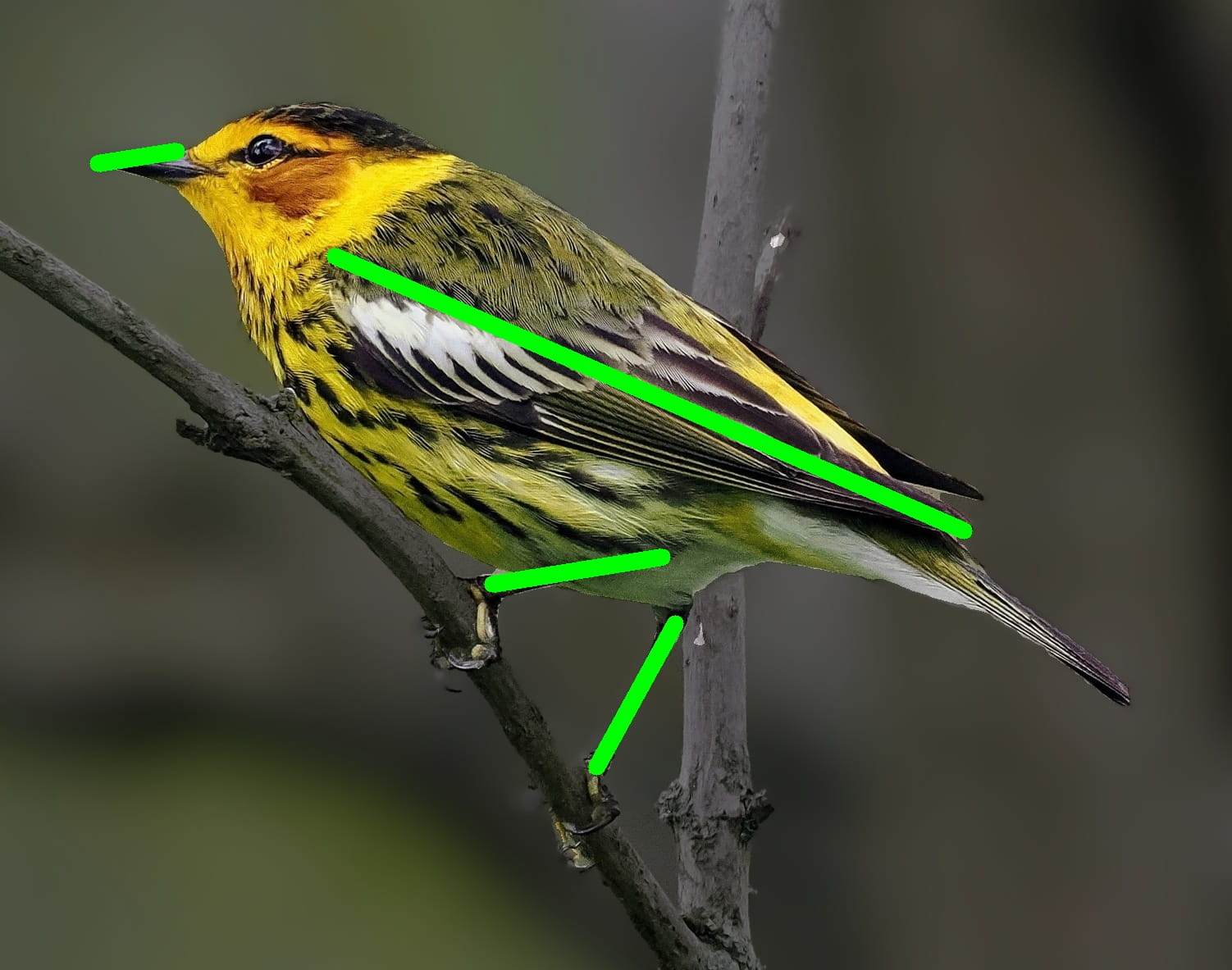}
        }
        \stackinset{r}{3pt}{b}{0pt}{\inatlabel{@erickolb}}{%
            \includegraphics[width=\linewidth]{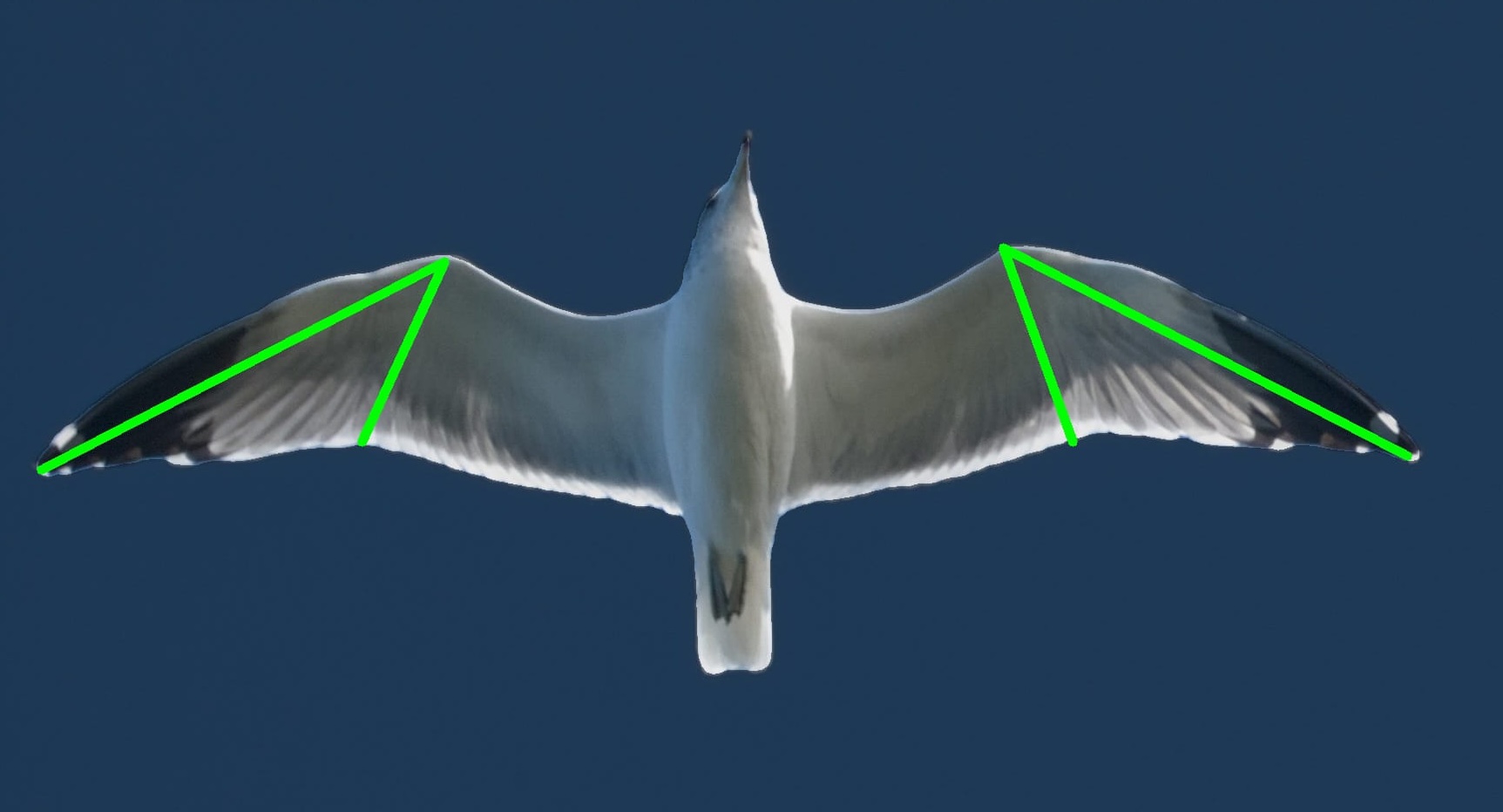}
        }
    \end{minipage}
    
\end{table}

\customparagraph{Shorebirds.}
\cref{tab:shore} shows results on five species from the Alaska Science Center adult shorebird dataset~\cite{shore}. Using a single reference image per species, we measure four parts: exposed culmen (an alternative measure of bill length), tarsus, head length, and wing length. 
On this dataset, our measurements have about 11\% relative error compared to the ground truth. We observe more accurate results for tarsus and culmen ratios than for head length. 
We speculate that keypoint matching is less accurate for head length because the back-of-head keypoint is less distinctive than the keypoints used for other parts, such as the bill tip or wing tip.

\customparagraph{Frogs.}
\cref{tab:frogs} shows results on five frog species covered in the University of Michigan Deep Blue Data~\cite{frogs}.
Using two reference images per species, we measure four distances: eye-eye, eye-nostril, head width, and nostril-nostril. 
Similar to the head length for shorebirds, we attribute the poor head width estimate to indistinct keypoints for this measure compared to other parts. 
Despite the large number of available images, the canonical poses required for our analysis were relatively rare for the Pacific 
Chorus Frog, which likely explains its worse performance across parts.

\begin{table}[t]
    \centering
    \setlength{\tabcolsep}{3pt}
    \caption{\textbf{\shore~\cite{shore}.} Relative error (\%) in median proportions. 
    To the right is an example query: Ruddy turnstone with culmen, head, tarsus, and wing keypoints.}
    \label{tab:shore}
    \vspace{-2mm}
    \begin{minipage}{0.74\textwidth}
    \begin{tabular}{lcccc}
    \toprule
    \multirow{2}{*}{Species} & Tarsus  & Culmen & Total Head  & \multirow{2}{*}{Mean}\\
    & / Wing L. & / Wing L. & / Wing L. \\
    \midrule
    Ruddy Turnstone & 5.2 & 7.3 & 13.9 & 8.8  \\
    Dunlin & 12.8 & 7.5 & 9.0 & 9.8  \\
    Western Sandpiper & 15.4 & 15.3 & 23.1 & 17.9  \\
    Bar-tailed Godwit & 11.0 & 5.9 & 17.6 & 11.5  \\
    Lesser Yellowlegs & 5.4 & 6.8 & 9.0 & 7.1  \\
    \midrule
    Mean & 10.0 & 8.6 & 14.5 & 11.0  \\
    \bottomrule
    \end{tabular}
    \end{minipage}
    \begin{minipage}{0.24\textwidth}
    \stackinset{r}{3pt}{b}{0pt}{\inatlabel{@etienne\_gaillard}}{%
        \includegraphics[width=\linewidth]{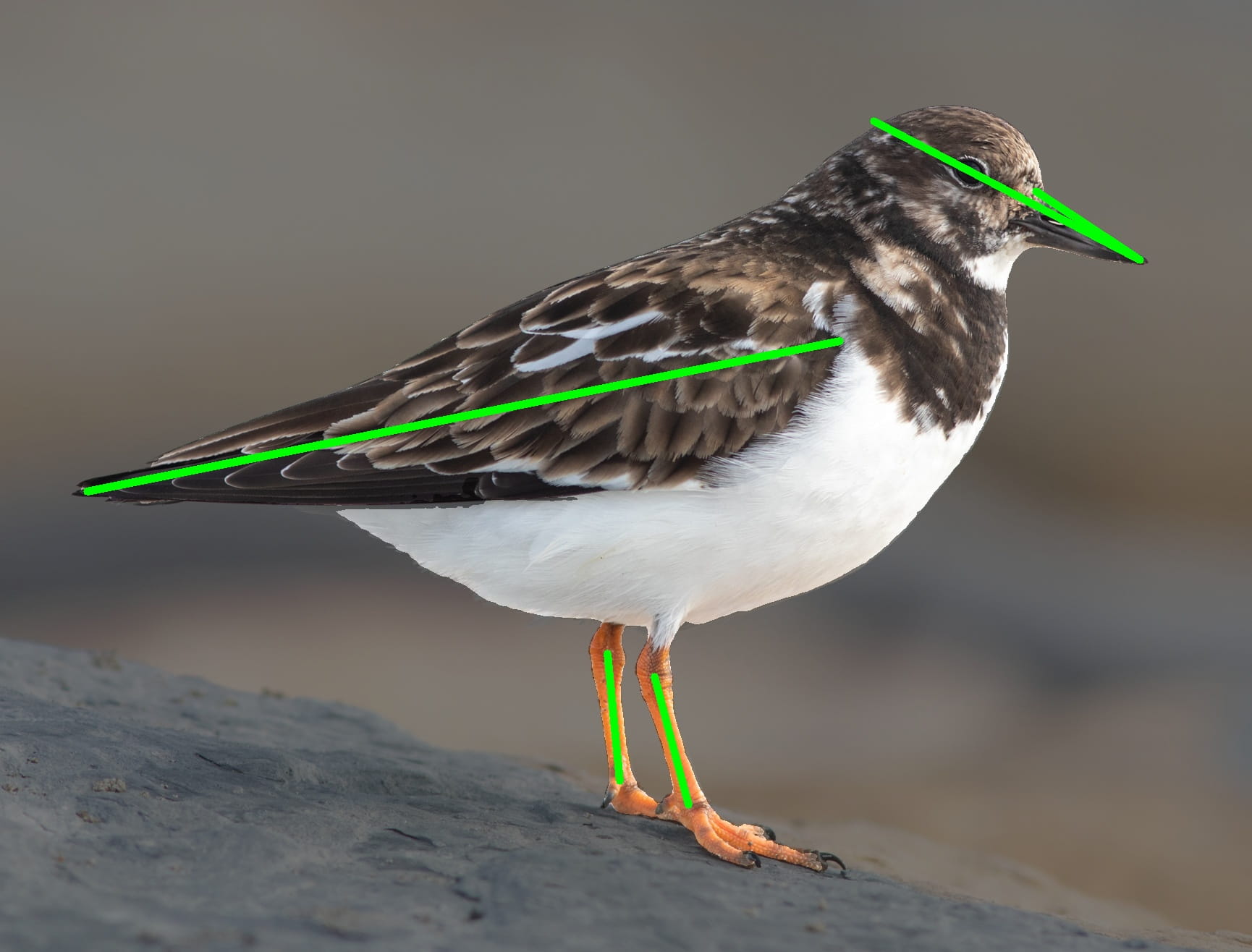}
    }
    \end{minipage}
\end{table}

\customparagraph{Discussion of quantitative results.}
We do not observe a consistent pattern across all species or parts, as estimation error is influenced by the abundance of target poses, the distinctiveness of the keypoints, and the size of the part relative to the image. 
Barring a few exceptions, the estimated proportions are within 10–20\% of the actual physical measurements, and inter-species trends align with expectations (\eg longer tarsus for Egret than Warbler  in~\cref{fig:quantiles}).

\begin{table}[t]
    \centering
    \setlength{\tabcolsep}{1.5pt}
    \caption{\textbf{\frogs~\cite{frogs}.} Relative error (\%) in median proportions. Lower is better. To the right are example query images and keypoints: front view for Emerald Glass Frog (eyes, nostrils, and head width); and top view for Red-eyed Tree Frog (eyes and nostrils).}
    \label{tab:frogs}
    \vspace{-2mm}
    \begin{minipage}{0.79\textwidth}
    \begin{tabular}{lcccc}
    \toprule
    \multirow{2}{*}{Species} & Eye-Nostril   & Head Width  & Nostril-Nostril  & \multirow{2}{*}{Mean}\\
    & / Eye-Eye & / Eye-Eye & / Eye-Eye \\
    \midrule
    Red eyed Tree Frog & 18.9 & 12.6 & 12.6 & 14.7 \\
    Woodhouse's Toad & 8.3 & 30.0 & 18.5 & 18.9 \\
    Emerald Glass Frog & 12.6 & 21.2 & 11.4 & 15.1 \\
    Cuban Tree Frog & 7.5 & 22.8 & 17.1 & 15.8 \\
    Pacific Chorus Frog & 34.4 & 22.3 & 25.3 & 27.3 \\
    \midrule
    Mean & 16.3 & 21.8 & 17.0 & 18.4 \\
    \bottomrule
    \end{tabular}
    \end{minipage}
    \begin{minipage}{0.18\textwidth}
    \stackinset{r}{3pt}{b}{0pt}{\inatlabel{@floris\_heemskerk}}{%
        \includegraphics[width=\linewidth]{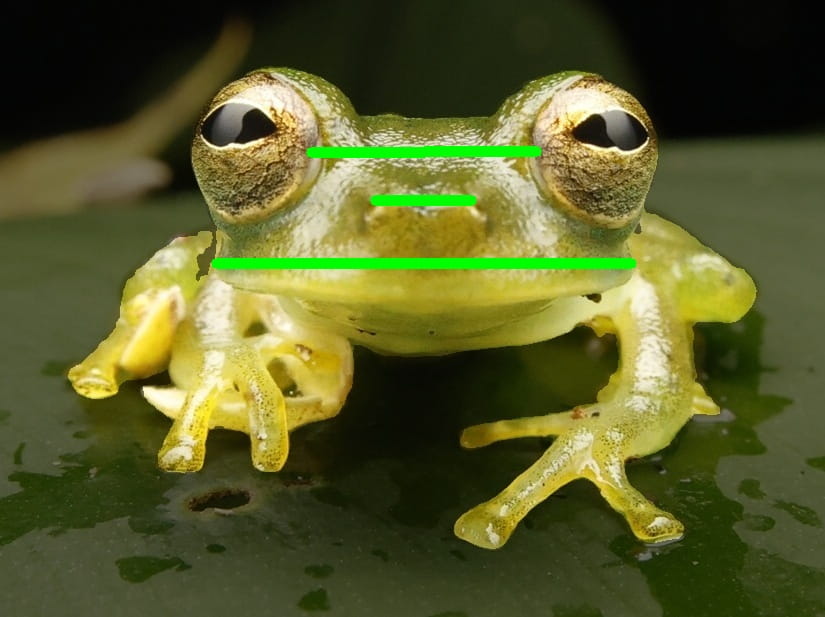}
    }
    \stackinset{r}{3pt}{b}{0pt}{\inatlabel{@amypi}}{%
        \includegraphics[width=\linewidth]{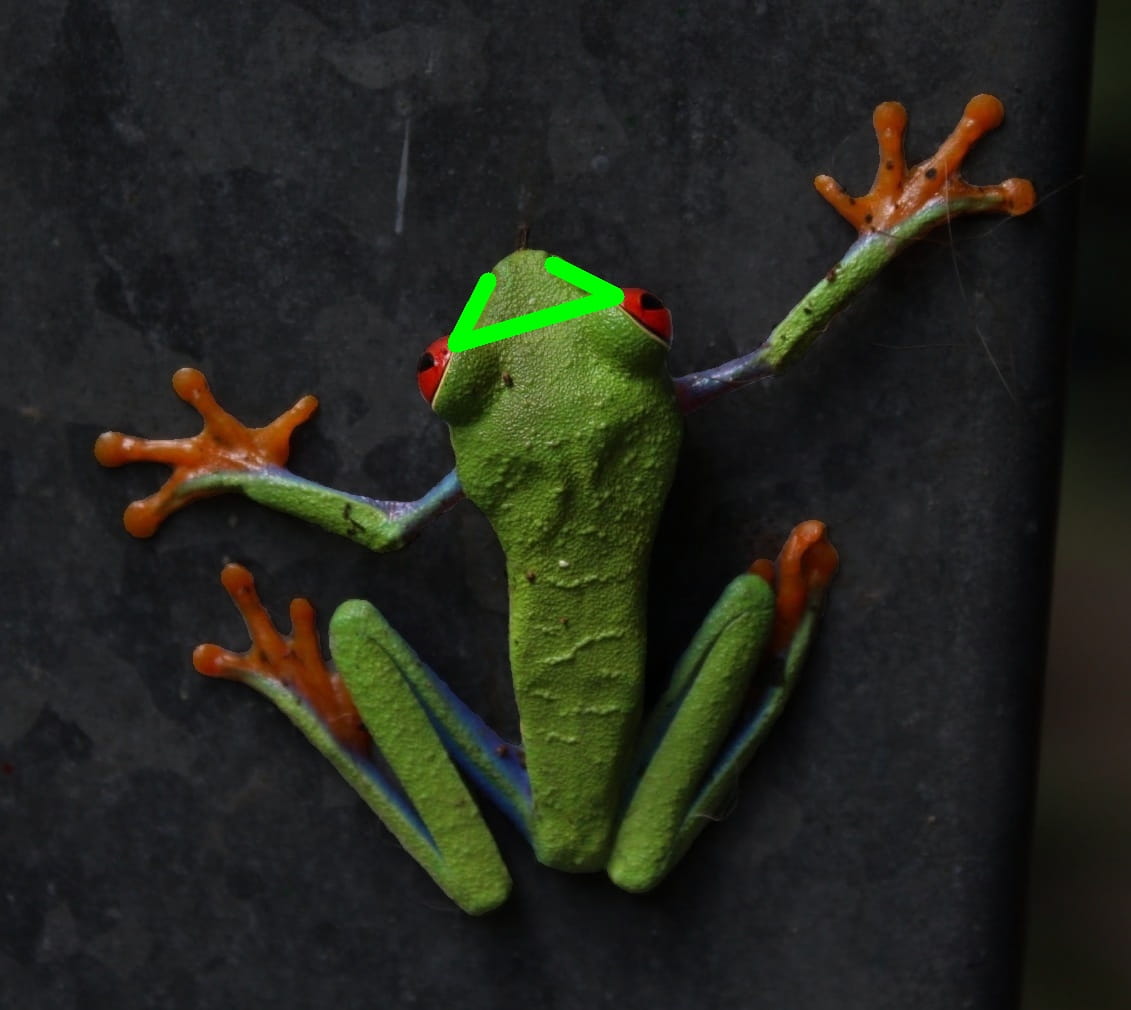}
    }
    \end{minipage}
\end{table}

\begin{wrapfigure}{r}{0.4\textwidth}
  \centering
  \includegraphics[width=0.98\linewidth]{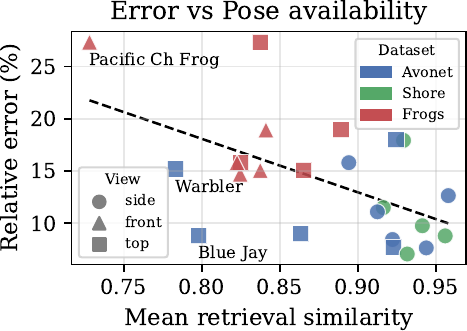} 
    \vspace{-4mm}
  \caption{\textbf{Pose availability. } Relative error vs mean retrieval similarity for each species and pose.}
  \label{fig:pose_availability}
  \vspace{-6mm}
\end{wrapfigure}
\customparagraph{Pose availability \& model performance.} The availability of images in suitable poses is an important factor influencing \method performance and likely explains several trends observed in our experiments (\eg, Pacific Chorus Frog in the front-view setting, and Warbler and Blue Jay in the bottom-view setting). To investigate this effect, we quantify pose availability using the mean similarity of the top 100 retrieved images and compare it with estimation error in~\cref{fig:pose_availability}. The observed negative correlation suggests that retrieval similarity can serve as a useful predictor of performance: when more pose-aligned images are available, \method tends to produce more accurate estimates.

\begin{figure}[t]
    \centering
    \includegraphics[width=0.96\linewidth]{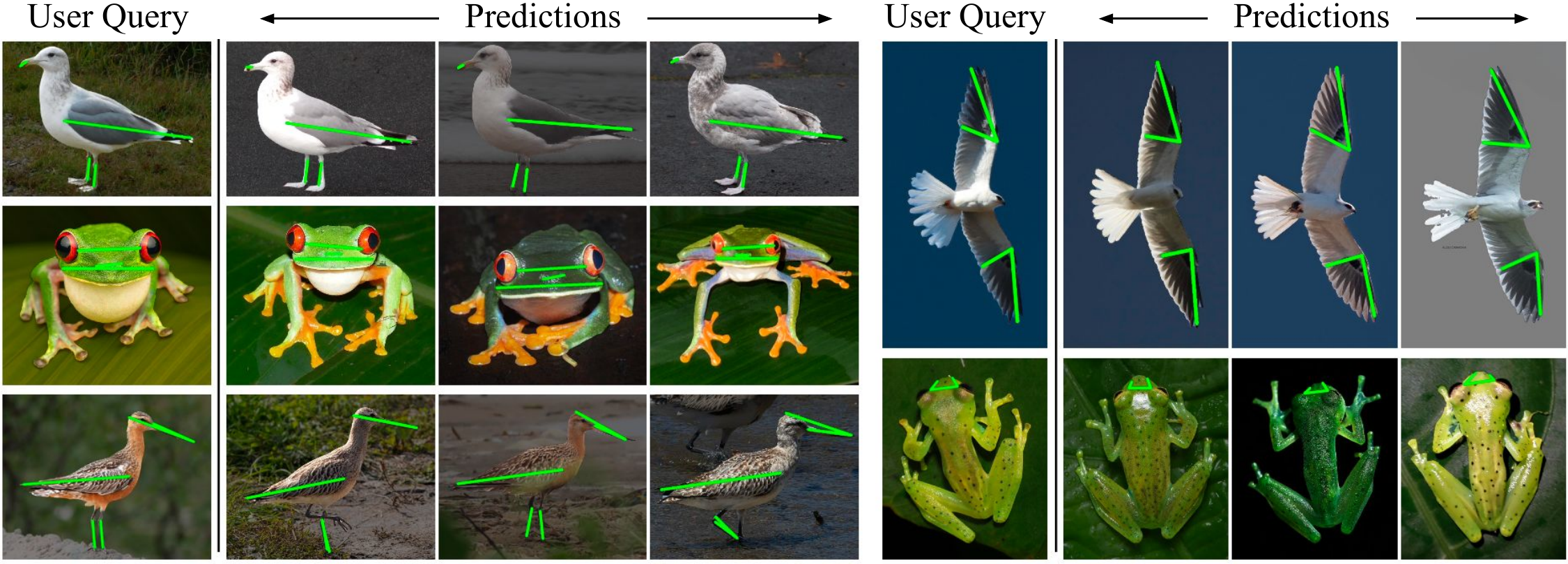}
    \vspace{-1mm}
    \caption{\textbf{Prediction Visualizations.} Sample queries and model predictions. We demonstrate the pose-similarity of retrieved images, and matched keypoints on them. }
    \label{fig:pred_vis}
    \vspace{-3mm}
\end{figure}

\subsection{Ablations}
\label{sec:ablation}

We ablate the key components of \method in \cref{tab:ablations}. 
First, we explore the contribution of each module in the overall pipeline (\cref{tab:abl:comp}). 
Using estimates from the query image alone results in a 12.2\% \avonet relative error. However, \method also provides an estimate of the full distribution, capturing intra-species variation, and supporting cross-species queries (\eg birds of prey in~\cref{fig:case_studies}).
Matching keypoints on randomly sampled images yields an exorbitant 216\% error, highlighting the difficulty of measuring parts from arbitrary poses. Restricting our analysis to pose-retrieved images reduces the errors to  40.0\%. Using background-suppressed images 
based on SAM~\cite{sam} segmentation reduces the error to 16.1\%, while RANSAC-based filtering further reduces it to 12.2\%. Finally, iterative refinement leads to a modest 1\% drop in error.

Next, we compare different model variants for retrieval (\cref{tab:abl:retrieval}) as well as for keypoint matching and iterative refinement (\cref{tab:abl:matching}).
CLIP-based models used for retrieval instead of DINO lead to worse performance, underscoring the importance of features that encode geometry and pose. Among the DINO variants, using the CLS token yields roughly the same performance as using a concatenation of all patch tokens, while switching to DINOv2~\cite{oquab2023dinov2} improves \avonet performance to 10.1\% relative error but worsens validation performance (hence we use DINOv3 instead). 
For keypoint matching, using a simple unweighted average leads to a modest increase of 0.7\%, and removing the restriction to top-$k$ retrieved images during aggregation does not affect the relative error on \avonet. Among the model variants, using DINOv2 features for keypoint matching yields a 2.2\% higher error. Interestingly, while Stable Diffusion~\cite{esser2024scaling} features worsen performance to 25.2\% relative error, concatenating DINOv3 and Stable Diffusion features performs best, with an 11.2\% error.

\begin{figure*}[t]
    \centering
\begin{subfigure}[!b]{0.32\linewidth}
    \includegraphics[width=\linewidth]{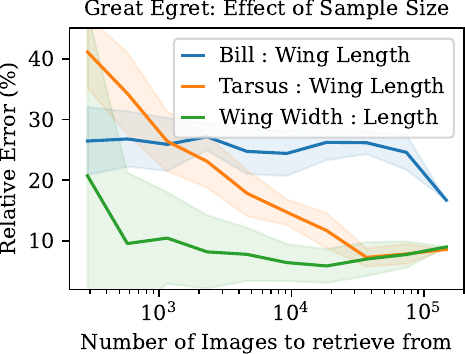}
    \subcaption{Effect of sample size}
    \label{tab:abl:size}
\end{subfigure}
\hfill
\begin{subfigure}[!b]{0.32\linewidth}
    \includegraphics[width=\linewidth]{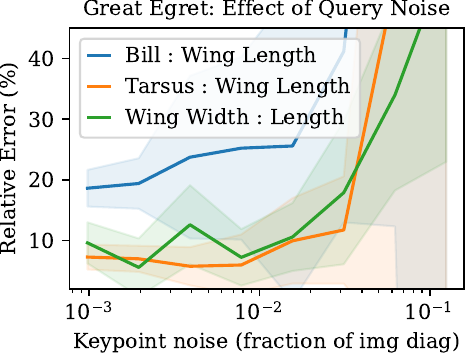}
    \subcaption{Effect of query noise}
    \label{tab:abl:noise}
\end{subfigure}
\hfill
\begin{subfigure}[!b]{0.32\linewidth}
    \includegraphics[width=\linewidth]{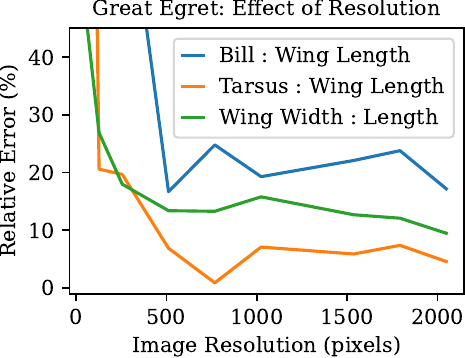}
    \subcaption{Effect of resolution}
    \label{tab:abl:resolution}
\end{subfigure}
    \vspace{-2mm}
    \caption{\textbf{Ablations.} We present errors for the Great Egret. (\textbf{a-b}): 10 runs mean $\pm$ std. }
    \label{fig:ablations}
    \vspace{-4mm}
\end{figure*}

\begin{figure}[!t]
    \centering
    \setlength{\tabcolsep}{6pt}
    \begin{tabular}{cc}
    \includegraphics[width=0.43\linewidth]
    {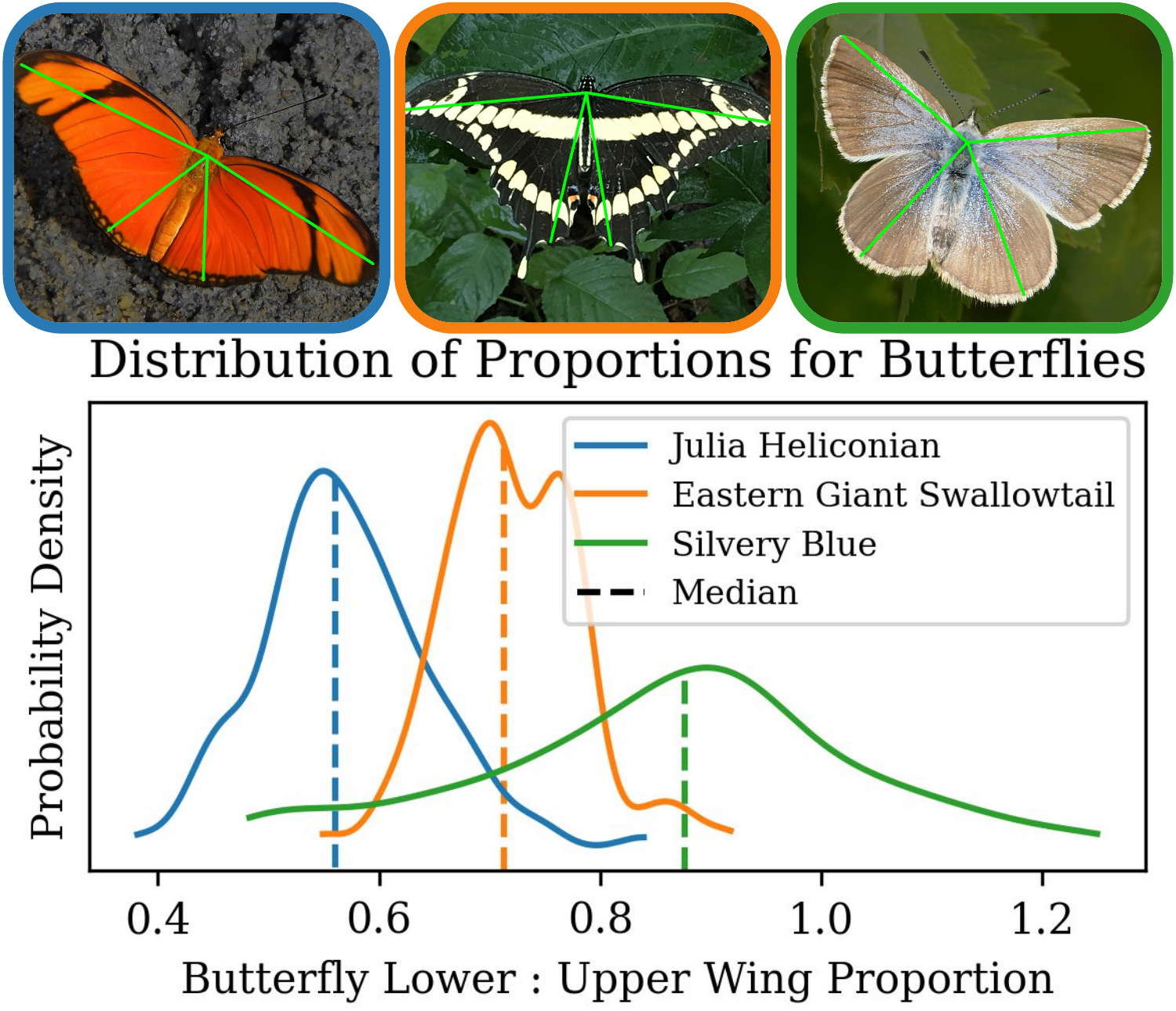} & 
    \includegraphics[width=0.43\linewidth]{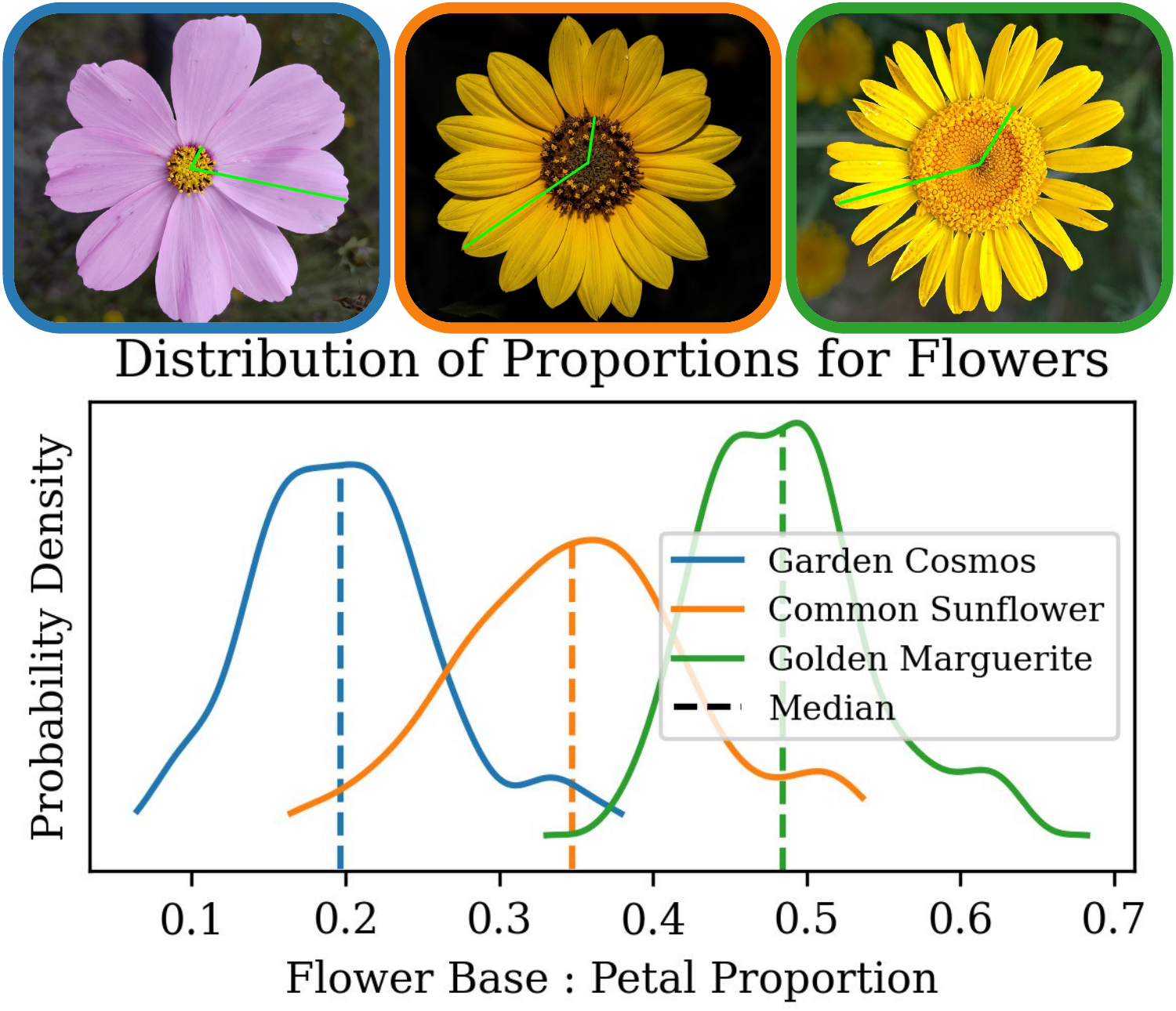} \\
    \includegraphics[width=0.43\linewidth]{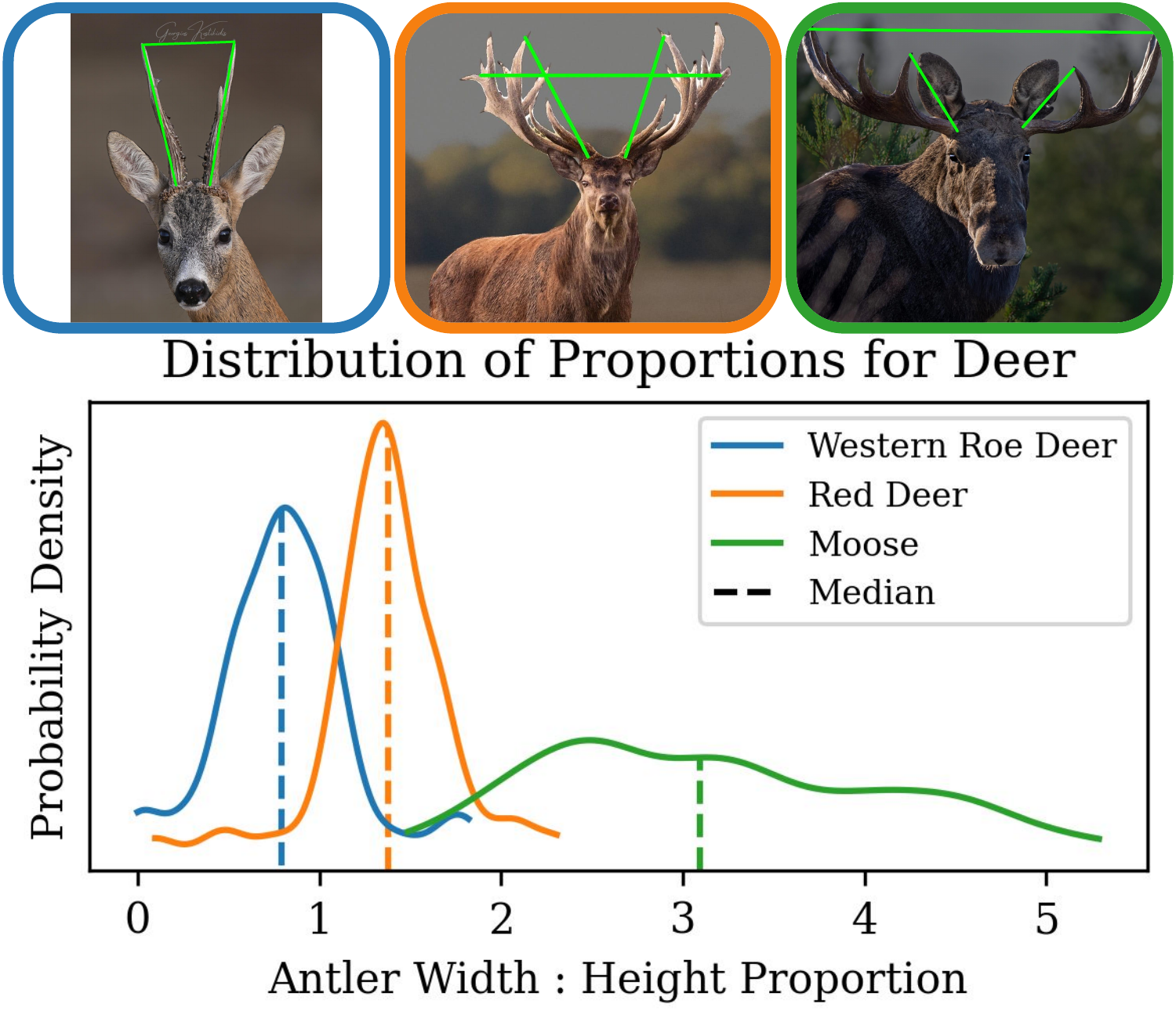} & 
    \includegraphics[width=0.43\linewidth]{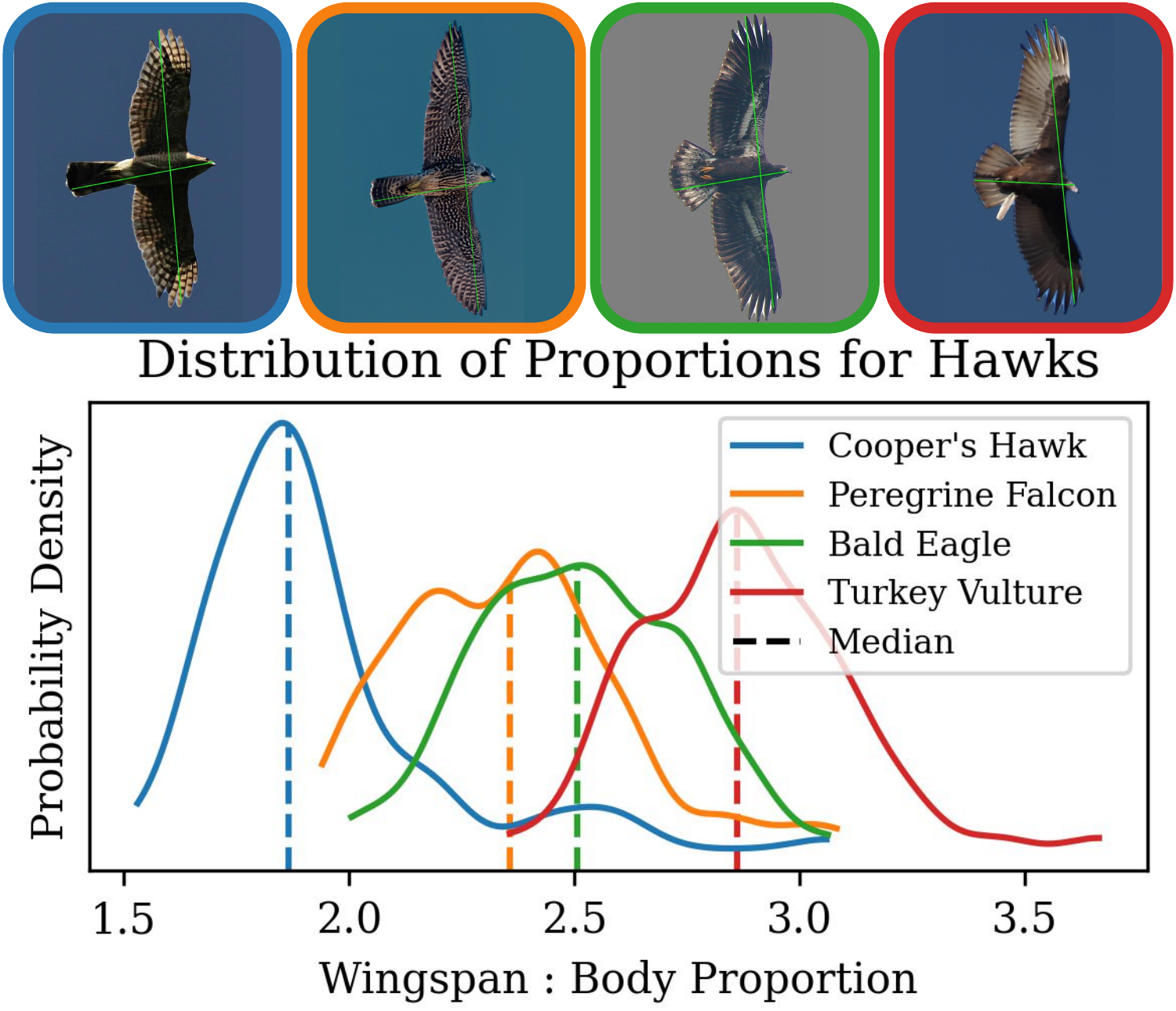} \\
    \end{tabular}
    \vspace{-3mm}
    \caption{\textbf{Case studies.} \method exhibits broad applicability across taxa and parts. Each panel here presents the estimated distribution of particular body proportions across a few species. Exemplar images and parts are shown for each case study.}
    \label{fig:case_studies}
    \vspace{-2mm}
\end{figure}

Next, we investigate the influence of retrieval sample size (\cref{tab:abl:size}) on estimation errors.
As the number of images available for a species increases, so does the likelihood of finding images with poses suitable for measurement, and consequently, measurement accuracy. We simulate different sample sizes by randomly sampling fractions of images and observe the resulting trend for the Great Egret, which has approximately 150K images. For smaller sample sizes, we observe retrieved images where parts of the legs are submerged in water, which likely explains the sharp increase in errors for the tarsus proportion. In contrast, the bill-to-wing-length proportion is more stable---as images with these parts clearly visible are abundant---and only begins improving near 100K images.

\begin{table*}[t]
    \caption{\textbf{Ablations.} For each experiment, we report relative errors in medians for \avonet and \colorbox{tablegray}{highlight} the row indicating the configurations used in \method. }
    \label{tab:ablations}
    \vspace{-2mm}
    \centering
\begin{subtable}[!b]{0.32\linewidth}
    \begin{tabular}{lc}
    \toprule
       Variant  & AVONET \\
       \midrule
       Query Only & 12.2 \\
       \midrule
        Random Samp. & 216.1\\
        + Retrieval & 40.0 \\
        + SAM & 16.1\\
        + RANSAC & 12.2\\
        \rowcolor{tablegray}+ Iterative & \textbf{11.3}\\
    \bottomrule
    \end{tabular}
    \subcaption{\method components}
    \label{tab:abl:comp}
\end{subtable}
\begin{subtable}[!b]{0.31\linewidth}
    \begin{tabular}{lc}
    \toprule
       Variant  & AVONET \\
    \midrule
        CLIP~\cite{radford2021learning} & 17.1 \\
        SigLIP~\cite{zhai2023sigmoid} & 11.8 \\
        BioCLIP~\cite{stevens2024bioclip} & 14.8 \\
        \midrule
        DINOv3-cls & 11.3 \\
        DINOv2-patch & \textbf{10.1}\\
        \rowcolor{tablegray}  DINOv3-patch & 11.3\\
    \bottomrule
    \end{tabular}
    \subcaption{Retrieval variants}
    \label{tab:abl:retrieval}
\end{subtable}
\begin{subtable}[!b]{0.35\linewidth}
    \begin{tabular}{lc}
    \toprule
       Variant  & AVONET \\
    \midrule
        \rowcolor{tablegray}
        DINOv3 & 11.3 \\
        w/o weigh. mean & 12.0 \\
        w/o top-k kp filter & 11.3 \\
        \midrule
        DINOv2 & 13.5 \\
        Stable Diff.~\cite{esser2024scaling} & 25.2\\
        DINOv3+SD~\cite{esser2024scaling} & \textbf{11.2} \\
    \bottomrule
    \end{tabular}
    \subcaption{Kp. Matching \& Refinement}
    \label{tab:abl:matching}
\end{subtable}
\vspace{-4mm}
\end{table*}

Next, we investigate the influence of query keypoint noise (\cref{tab:abl:noise}). Small errors in the query keypoints are natural for user-provided annotations and may arise from ambiguity in part endpoints (\eg the wing base covered by feathers), resolution constraints, or human error. We simulate such noise by adding Gaussian perturbations of varying relative magnitudes to the pixel coordinates of the query keypoints. For noise levels below 1\% of the image diagonal, the estimates remain stable for all three proportions. For context, in the query images, 1\% of the image diagonal corresponds to 14.9\%, 7.6\%, and 7.7\% of bill, tarsus, and wing-width, respectively. We observe sharp increases in measurement errors around 2–3\% noise, most notably for the tarsus-to-wing estimate, which increases from about 10\% to more than 40\% relative error. 

Lastly, we investigate the effect of image resolution (\cref{tab:abl:resolution}). We exclude RANSAC filtering for a clearer comparison, and vary image resolution only in the keypoint matching stage. At very low resolutions, parts may be compressed into a single patch, leading to a sharp increase in error. This is particularly evident for small parts such as the bill.
Higher resolutions tend to yield better results, with diminishing returns beyond 512.

\begin{figure}[t]
    \centering
    \includegraphics[width=0.495\linewidth]{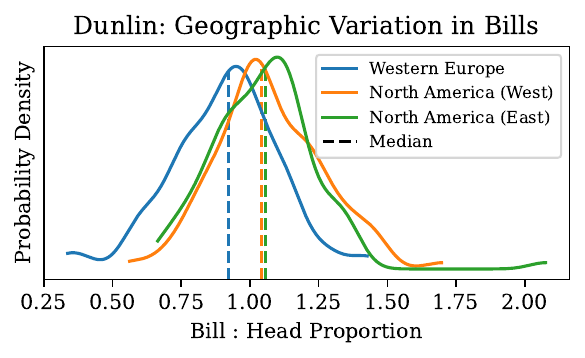}
    \includegraphics[width=0.48\linewidth]{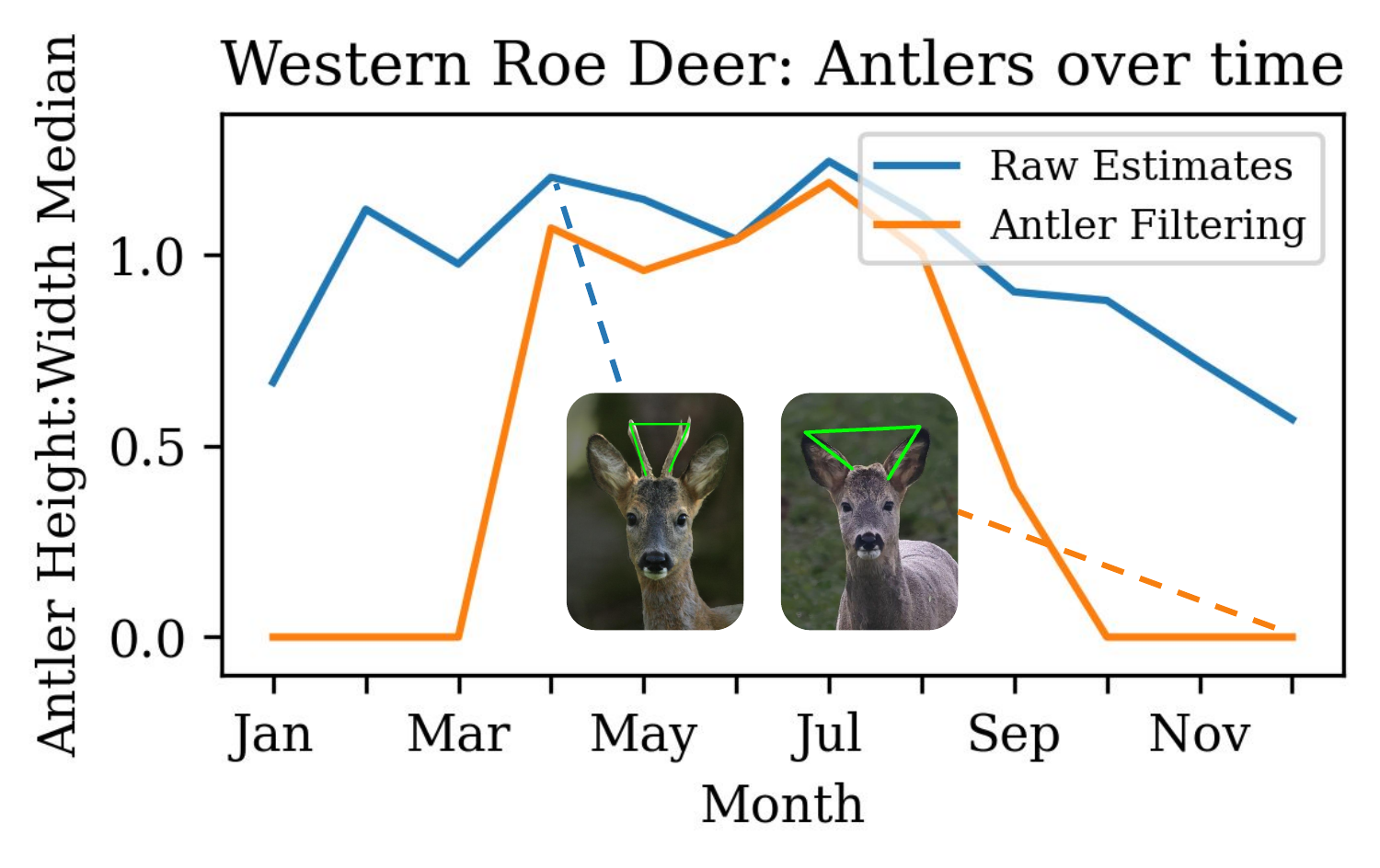}
    \vspace{-3mm}
    \caption{\textbf{Geographic and Temporal Stratification.} \method allows measurement of specific sub-populations based on criteria such as geographic location or time-of-year.}
    \label{fig:space_time}
    \vspace{-2mm}
\end{figure}

\subsection{Qualitative Case Studies}
\cref{fig:case_studies} explores the capabilities and limitations of \method through several qualitative case studies motivated by our interactions with ecologists and nature enthusiasts. Each case study focuses on a small set of species within the same taxonomic group and estimates body proportions that distinguish them. Additional case studies are presented in Appendix~A6.

\customparagraph{\method estimations distinguish species across diverse taxa.}
\method is broadly applicable to vastly different species and parts. 
We attempt measurement of upper and lower wing proportions for butterflies, base and petal proportions for flowers, and antler proportions for deer.
In each example, we find our measurements separate three distinct species within the taxonomic group and align with our expectations. 

\customparagraph{\method requires only a single query image for similar species.}
We analyze the ratio of wingspan to body proportion of various birds of prey in~\cref{fig:case_studies}.
In this case study, we use only a single query image for all species and find that pose and keypoint matching is fairly accurate across species.
Unlike most other families, birds of prey are frequently seen soaring overhead in harsh backlit conditions, requiring identification using only silhouette shape and flight style \cite{hawkidentification}. 
While identification can sometimes be possible using these proportions alone, we observe significant overlap in some species, particularly Peregrine Falcon and Bald Eagle. Other attributes which \method can not assess, such as absolute size and flight patterns, are likely required in these situations.

\customparagraph{\method can identify geographical and temporal variation.}
The search corpus in \method can be modified to obtain various stratified distributions, and \cref{fig:space_time} presents two such case studies.
First, we find \method is able to identify geographic variations in Dunlin. 
Wintering Dunlin populations in western Europe, and western and eastern North America represent three distinct subspecies populations: \textit{schinzii}, \textit{pacifica}, and \textit{hudsonia}, respectively. 
By stratifying our search corpora by location, we find the European population to have a median lower bill : head ratio than the North American populations, which are quite similar. 
This matches reported Dunlin subspecies studies, which find \textit{pacifica} and \textit{hudsonia} are the largest and longest-billed~\cite{botwdunlin}.
Although these studies indicate \textit{hudsonia} is slightly shorter-billed on average than \textit{pacifica}~\cite{botwdunlin}, we observe that \method is unable to distinguish these very similar subspecies.

Next, we examine seasonal change in antler length for Western Roe Deer.
Rather than at the species level, we filter images based on month and estimate measurements for each month separately.  
We observe the expected trend in antler heights when stags grow their antlers in early spring and shed them in October–December~\cite{kierdorf2023bone}.
Amusingly, \method selects ear tips as the closest match to the query antler tips when antlers are completely absent. Filtering out images that do not have antlers allows us to overcome this limitation and makes the trend even more apparent.

\subsection{Discussions and Limitations}

Rather than replacing physical measurement, we envision \method as a tool for rapid hypothesis generation and exploratory analysis before committing to more extensive measurement studies. While broad applicability across taxonomic groups and anatomical parts is a key strength of \method, it also comes with important limitations. First, estimates from \method are generally noisier than physical measurements, limiting its use in high-precision applications. Second, the image-based measurement paradigm has inherent constraints: scale ambiguity limits estimation to proportions rather than absolute metric lengths, and non-visual traits such as body mass or skeletal dimensions are not directly measurable from images alone. Third, because \method samples from citizen-science image collections, its estimates may reflect biases toward human population centers, observer preferences, visually distinctive species, and popular taxa.

Evaluation also remains challenging. Existing physical morphometric datasets are often small, unevenly sampled, or collected under protocols that differ from the available image data, making direct comparisons imperfect. In addition, certain viewpoints and poses can be difficult to photograph for some species (\eg bottom views or flight poses for small birds such as warblers and hummingbirds), limiting performance for proportions visible only under such conditions. Finally, beyond basic geometric heuristics, \method does not currently have a principled abstention mechanism. As a result, it may produce unrealistic keypoints or proportions when retrieval fails or when the specified body parts are absent.

\section{Conclusion and Future Work}
\label{sec:conclusion}

We present \method, a scalable approach for estimating wildlife body proportions from large, opportunistic image collections. Compared to physical measurement, \method provides a noisier but substantially more scalable alternative that can support ecological, biological, and evolutionary analyses when precise individual-level measurements are unavailable or impractical. Across three large-scale morphometric studies, our population-level body-proportion estimates are typically within 10--20\% of physical measurements. Our estimates also follow expected trends in several qualitative case studies spanning diverse taxonomic groups and body parts. In addition, \method enables measurements under different stratifications, supporting analyses such as geographic variation in Dunlin bill proportions and seasonal trends in deer antlers. Together, these results suggest that retrieval-conditioned correspondence offers a practical pathway for estimating geometric population statistics from opportunistic imagery.

Several directions remain for future work. While \method uses off-the-shelf foundation models for broad applicability, dedicated animal pose or correspondence models could improve robustness, especially for taxa, viewpoints, or body parts with limited pose-aligned imagery. Incorporating calibrated depth estimation or explicit 3D keypoint detection could further extend the approach to a broader range of poses and viewpoints, moving toward more reliable individual-level proportion estimates across large populations. Incorporating a small number of physical measurements, when available, could also help calibrate image-based estimates and partially alleviate scale ambiguity. Future work could explore richer anatomical representations, such as part segmentations, curves, or point clouds, which provide additional geometric context beyond endpoints. Finally, extending the framework to additional morphological traits, including inferred body mass, coloration, and shape descriptors, could broaden the scope of image-based morphometric analysis.

\paragraph{Acknowledgments.} The project was funded in part though grants \#2329927 and \#2504073 from the National Science Foundation, USA. We thank Feipeng Huang and Mark Titus for case study recommendations.

\bibliographystyle{splncs04}
\bibliography{main}

\clearpage

\appendix
\setcounter{table}{0}
\renewcommand{\thetable}{A\arabic{table}}
\setcounter{figure}{0}
\renewcommand{\thefigure}{A\arabic{figure}}

\section{Appendix}

We begin with additional implementation details and hyperparameter tuning in \cref{sec:apdx:implement}. 
We then report robustness experiments evaluating sensitivity to query pose in \cref{sec:apdx:pose_robust} and to query individual in \cref{sec:apdx:individual_robust}. 
Additional comparisons of distribution statistics are presented in \cref{sec:apdx:violin}.
Prediction visualizations for the quantitative experiments, highlighting both success and failure cases, are provided in \cref{sec:apdx:pred_vis}. 

We next present an additional case study and two additional geographic stratification experiments in \cref{sec:apdx:case_studies}. 
To illustrate the challenges of measuring body parts from arbitrary images in uncurated collections, we show several California Gull images from iNaturalist and describe why each is not measurable in \cref{sec:apdx:uncurated}. 
Finally, we include additional prediction visualizations for the qualitative case studies and spatiotemporal stratification experiments in \cref{sec:apdx:vis_case_studies}.


\begin{figure}[ht]
    \centering
    \includegraphics[width=0.49\linewidth]{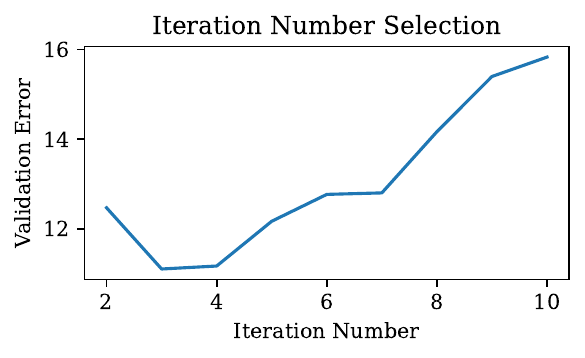}
    \includegraphics[width=0.49\linewidth]{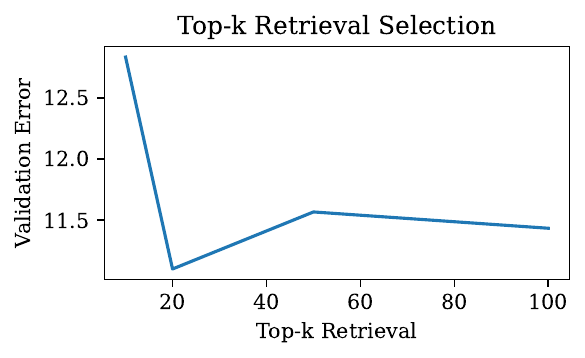}
    \includegraphics[width=0.49\linewidth]{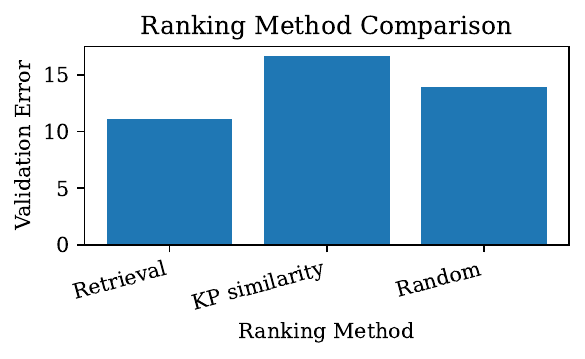}
    \includegraphics[width=0.49\linewidth]{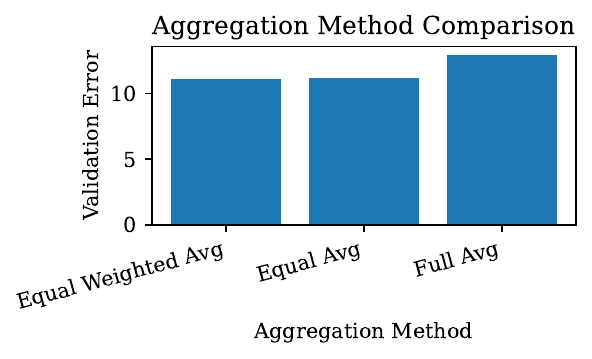}
    \caption{\textbf{Hyperparameter estimation.} Median relative errors, averaged across different parts for the validation set. }
    \label{fig:apdx:hyperparameter}
\end{figure}

\subsection{Additional Implementation Details}
\label{sec:apdx:implement}
In~\cref{fig:apdx:hyperparameter}, we show validation errors for different hyperparameters of the iterative refinement step; lower values indicate better performance. Three to four refinement iterations perform best, while additional iterations degrade performance. For selecting the top-$K$ keypoints used for aggregation, $K=20$ substantially improves over $K=10$, but performance deteriorates when using larger values of $K$. 

We also compare different ranking criteria for selecting the top-$K$ keypoints. Surprisingly, ranking by the original retrieval order performs better than ranking by keypoint similarity, suggesting that global image similarity may provide more informative and diverse keypoint features for refinement. Even random ordering outperforms keypoint-similarity ranking, indicating that selecting only the most similar keypoints may reduce the diversity needed for robust adaptation. Finally, we evaluate several aggregation variants for the selected keypoint features. The best performance is obtained by combining a similarity-weighted average of the retrieved keypoint features with the original query keypoint feature using equal weights.

\begin{figure*}[t]
\begin{minipage}{0.55\linewidth}
\centering
     \includegraphics[height=48mm]{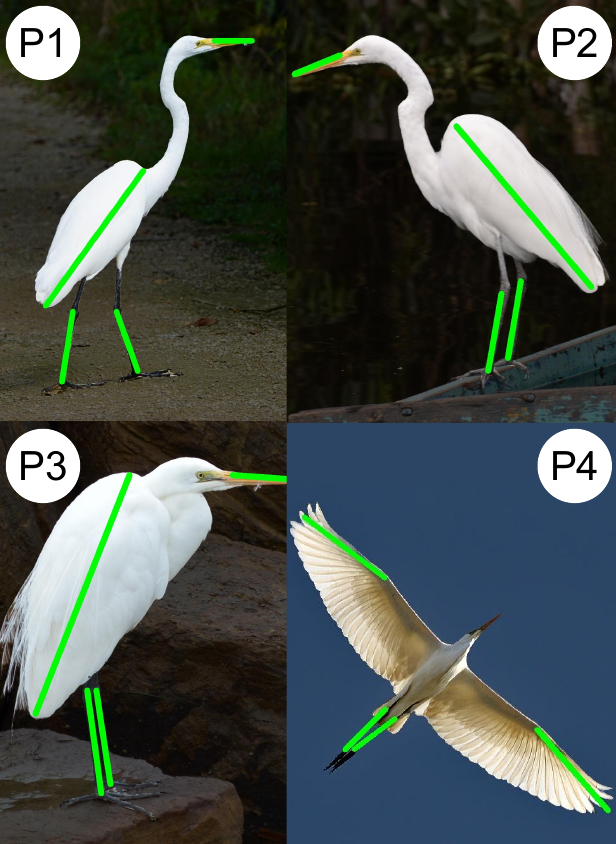}
      \includegraphics[height=48mm]{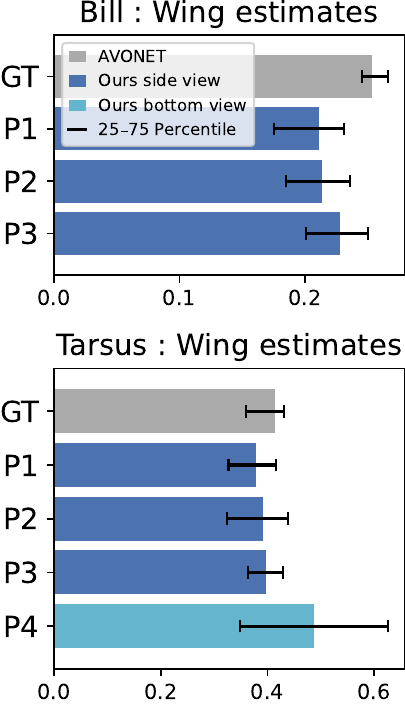}
\caption{Robustness to query pose.}
\label{fig:apdx:query_pose}
\end{minipage}
\hfill
\begin{minipage}{0.44\linewidth}
\centering
    \includegraphics[height=48mm]{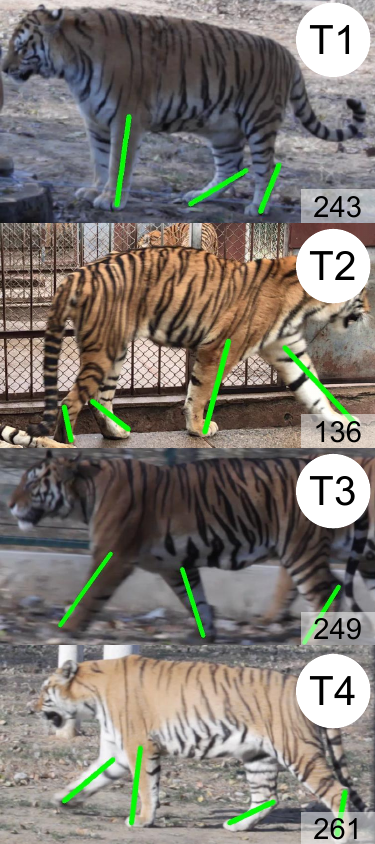}
    \includegraphics[height=48mm]{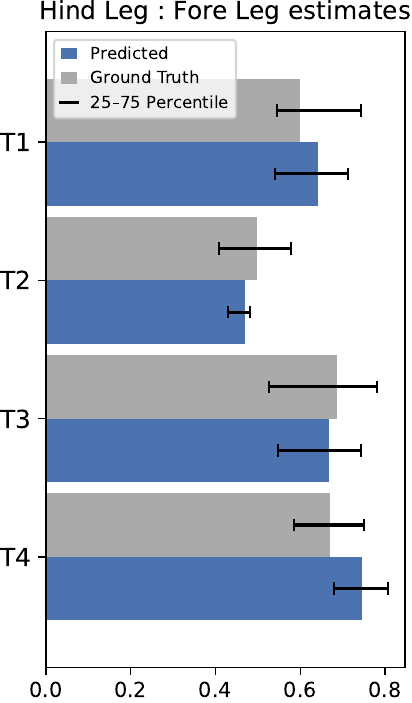}
\caption{Individual variation.}
\label{fig:apdx:tiger}
\end{minipage}
\end{figure*}
\subsection{Robustness to Query Pose}
\label{sec:apdx:pose_robust}
\method takes a single query image as input and retrieves images with similar poses. However, the same body proportion can often be measured from several different query poses, each of which may be valid. In this section, we investigate the robustness of \method to the choice of query pose.

We compare predicted proportions, summarized by medians and quartiles, from four different poses of the Great Egret in~\cref{fig:apdx:query_pose}. For pose variations within the general side-facing view (P1--P3), the estimates are relatively consistent and broadly fall within one another's interquartile ranges. The tarsus ratio can also be estimated from the bottom view, although we observe higher variation in this setting, likely due to the relative rarity of such images. We envision our approach as part of a human-in-the-loop workflow in which experts select appropriate query images and inspect retrieved images and estimated keypoints to assess the reliability of the resulting estimates.

\subsection{Individual Variation}
\label{sec:apdx:individual_robust}
With \method, our goal is to capture variation across individuals of the same species. However, even images of the same individual may yield different estimates due to foreshortening, pose changes, or keypoint matching errors. In this section, we investigate how \method estimates vary across different images of the same individual.

We experiment with four tigers from the ATRW animal re-identification dataset~\cite{li2020atrw}. For each individual, we use the annotated keypoints from one image as the query and estimate proportions in other images of the same individual. The selected tigers, along with their IDs and results, are shown in~\cref{fig:apdx:tiger}. The estimated proportions are generally close to the ground truth, and their variation is comparable to labeling noise, measured as the variation in proportions derived from manually labeled keypoints.

\begin{figure}
    \centering
    \includegraphics[width=\linewidth]{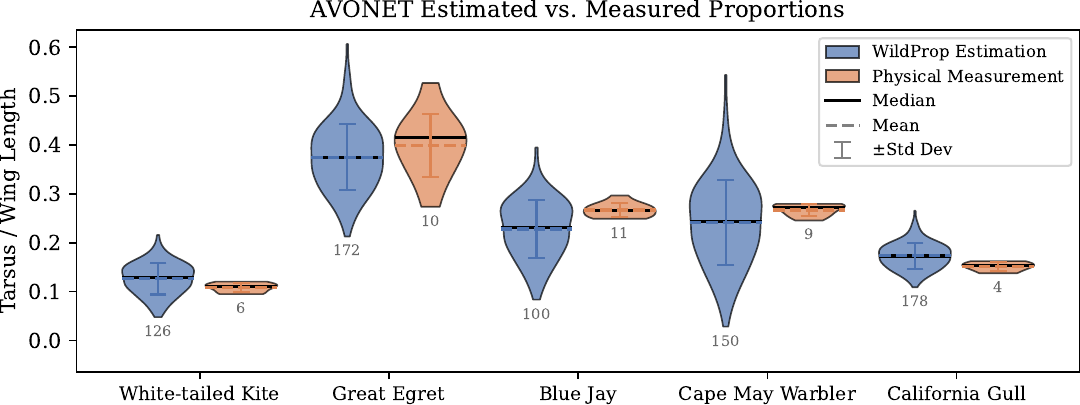}
    \newline
    \newline
    \includegraphics[width=\linewidth]{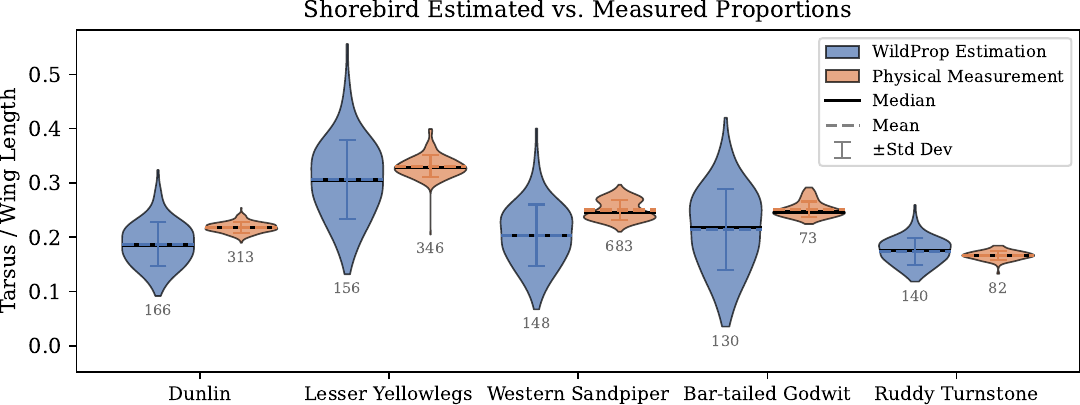}
    \newline
    \newline
    \includegraphics[width=\linewidth]{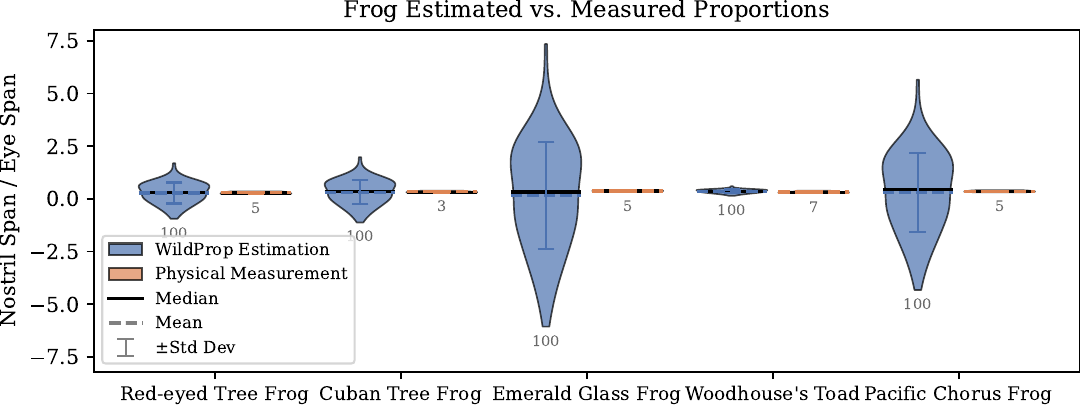}
    \caption{\textbf{Comparison of \method estimates and physical measurements for the \avonet, \shore, and \frogs datasets.} For each species, we show the estimated and measured proportion distributions, along with their mean, median, and standard deviation.}
    \label{fig:apdx:violin}
\end{figure}

\subsection{Comparing Additional Distribution Statistics}
\label{sec:apdx:violin}
In addition to the medians  and quartiles shown in the main paper, we compare additional statistics of the proportion distributions estimated by \method and those measured physically in \avonet, \shore, and \frogs.
Results are shown in~\cref{fig:apdx:violin}. Overall, the medians and means follow similar trends, while the larger standard deviations of \method estimates reflect sensitivity to outlier predictions.

\begin{figure}
    \centering
    \includegraphics[width=\linewidth]{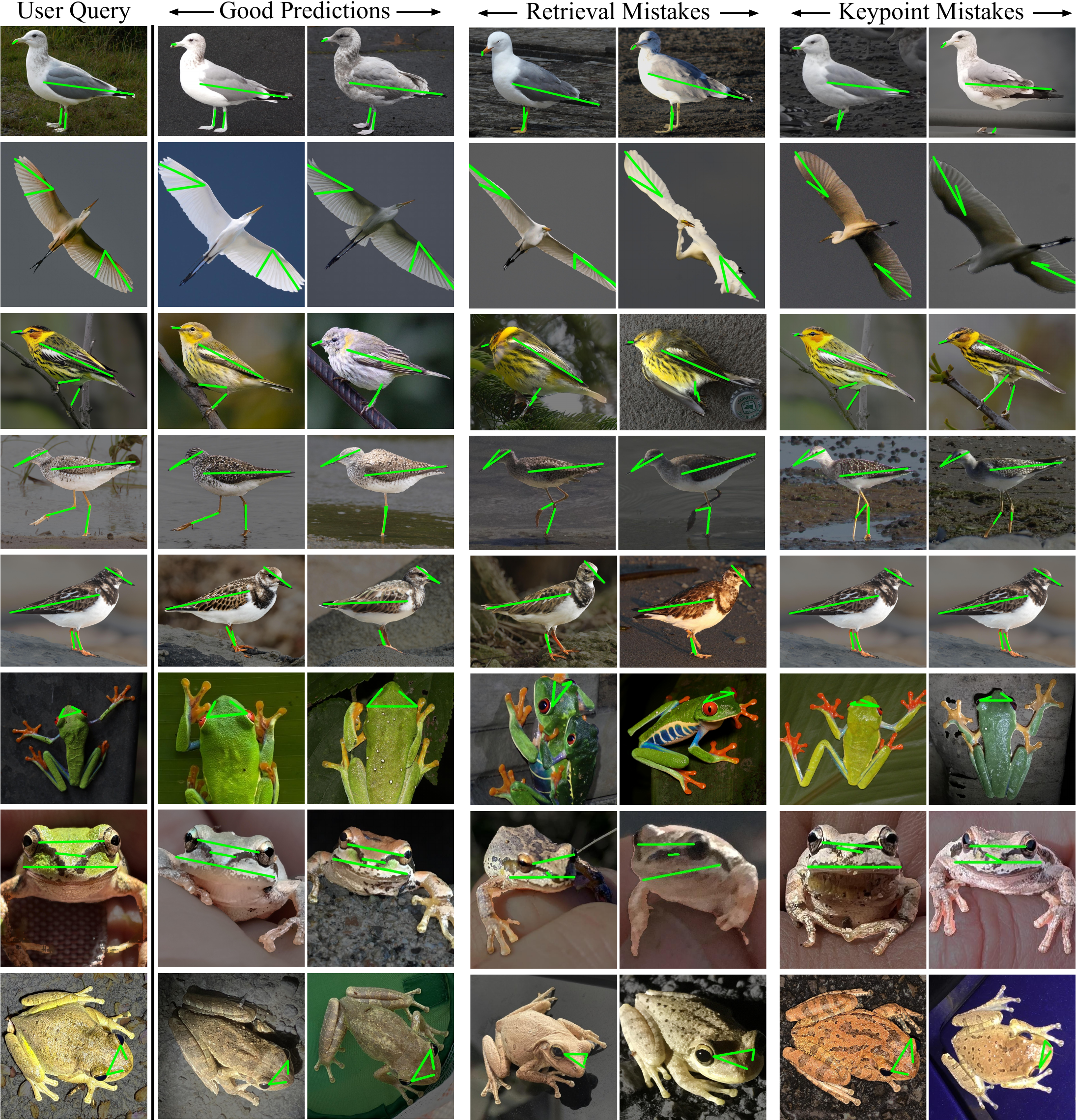}
    \caption{\textbf{Prediction visualizations for quantitative studies.} Each row shows the user query with annotated keypoints, followed by six sample predictions. We show two successful predictions, two examples where retrieval selects images in slightly or substantially different poses that affect measurement, and two examples where the pose is appropriate but keypoint detection fails.}
    \label{fig:apdx:pred_vis}
\end{figure}

\subsection{Additional Prediction Visualizations}
\label{sec:apdx:pred_vis}
We present prediction visualizations for the species used in our quantitative experiments (\avonet, \shore, and \frogs) in~\cref{fig:apdx:pred_vis}. Each row shows a query image in a specific pose, with the  parts to be measured indicated by green lines. We then show several successful predictions, as well as representative failure cases caused by errors in either pose-aware retrieval or keypoint detection. These examples illustrate both the reliability of \method when suitable pose-aligned images are retrieved and its failure modes when retrieval or correspondence breaks down.

\subsection{Additional Case Studies}
\label{sec:apdx:case_studies}

\begin{figure}[ht]
    \centering
    \includegraphics[width=0.49\linewidth]{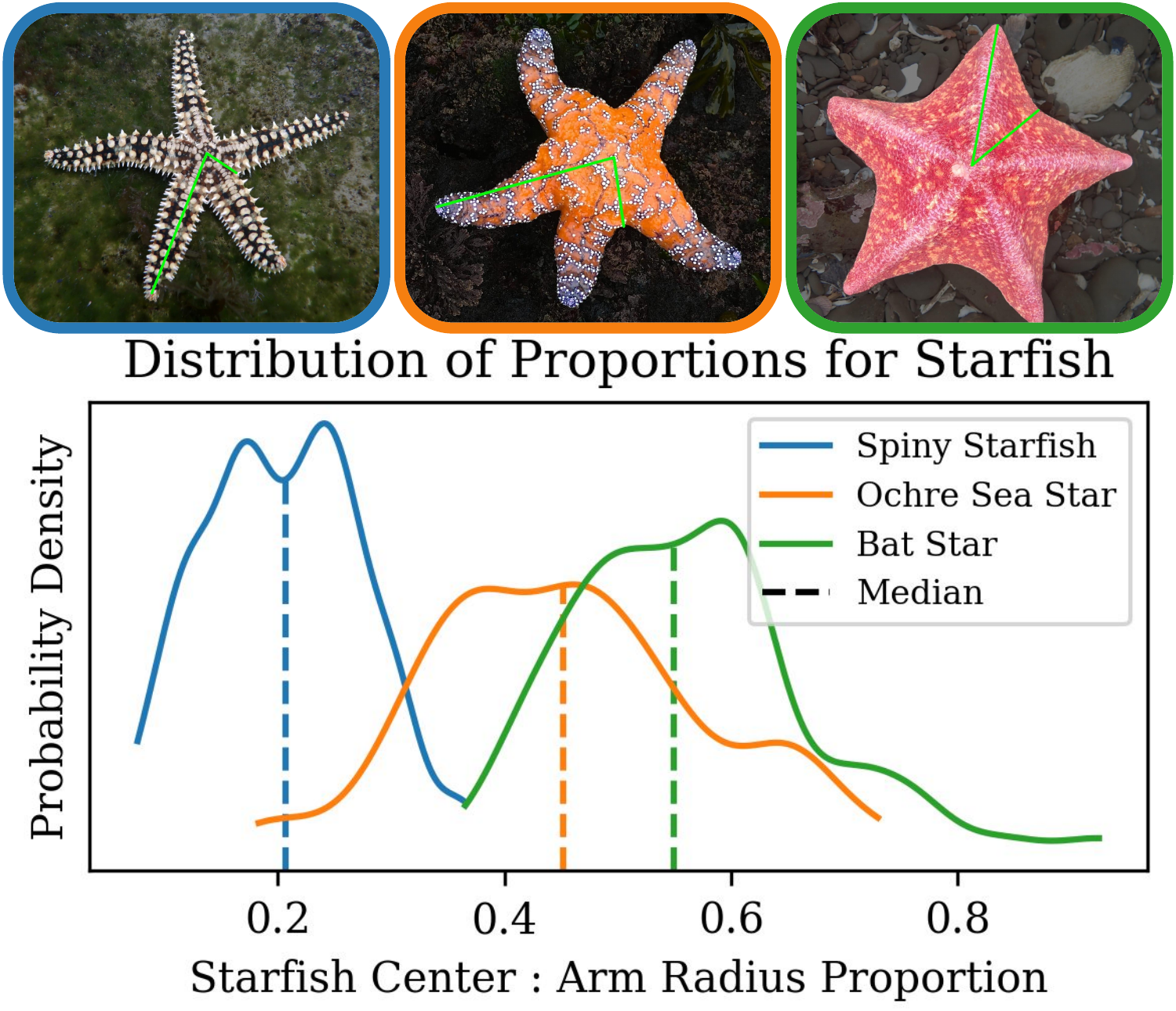} 
    \includegraphics[width=0.49\linewidth]{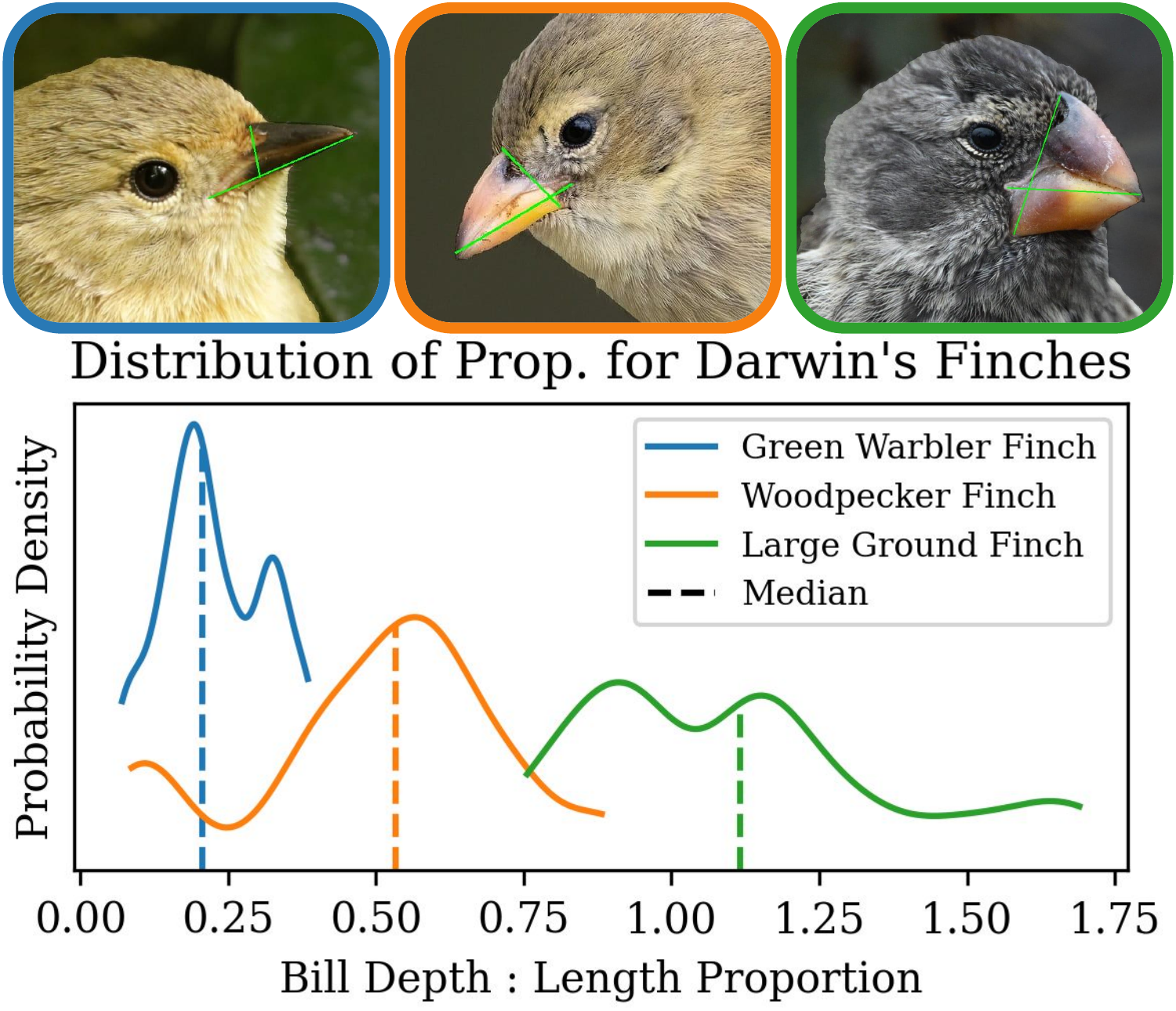}
    \includegraphics[width=0.49\linewidth]{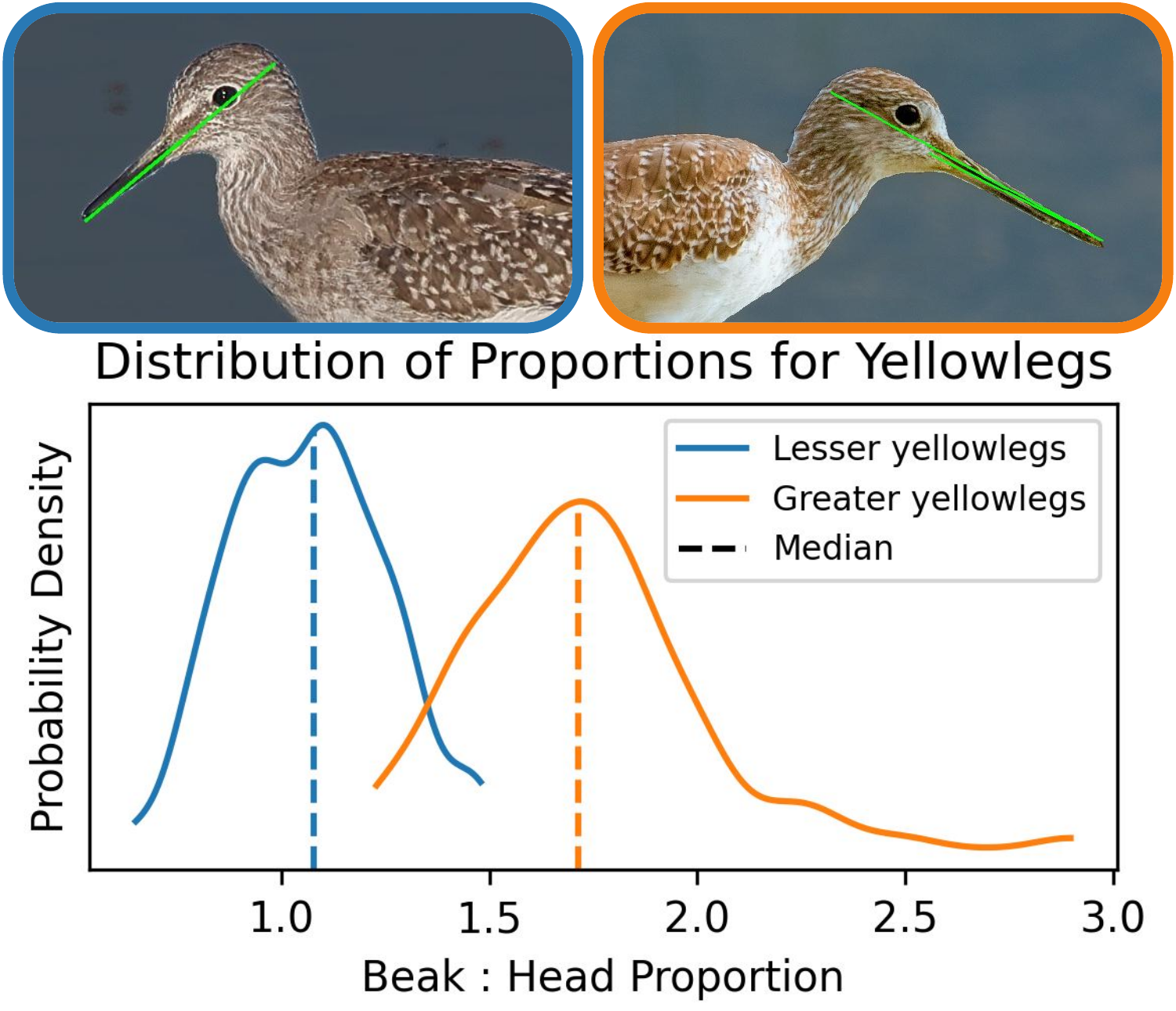}
    \caption{\textbf{Additional case studies.} We study the difference between 1) center to arm proportions for starfish, 2) bill length to depth proportions for Darwin's finches, and 3) the bill proportions of lesser and greater yellowlegs.}
    \label{fig:apdx:case_studies}
\end{figure}

\method is applicable to a wide range of species. In this section, we present additional case studies (Figure~\ref{fig:apdx:case_studies}) similar to the ones in the main paper, Fig~8. First are starfish, where the center to arm proportions separate three distinct starfish species. The bill shapes of Darwin's finches were key evidence used for his work on the theory of evolution. We observe that the bill depth to length estimates from \method are able to separate three species in Darwin's finches.

Greater and Lesser Yellowlegs are two similar shorebird species with similar appearance in all plumages, but can be distinguished by overall size and bill shape and size \cite{botwyellowlegs}.
Though overall size cannot be determined without direct comparison or in-hand measurements, bill size can be measured relative to head length.
We confirm this distinguishing feature can be identified in our system in Figure~\ref{fig:apdx:case_studies}---the Yellowlegs species have almost no overlap in relative bill length as measured by \method.

\begin{figure}[t]
    \centering
    \includegraphics[width=0.49\linewidth]{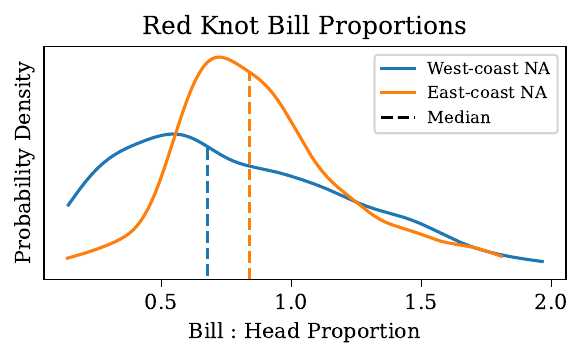}
    \includegraphics[width=0.49\linewidth]{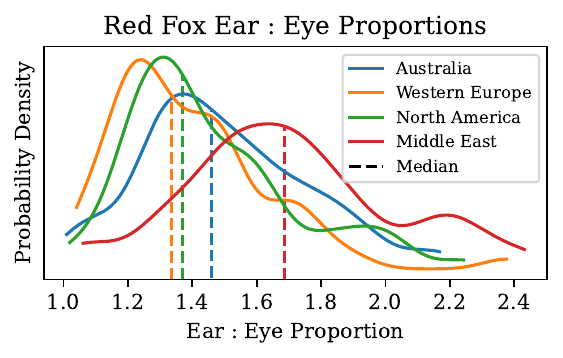}
    \caption{\textbf{Geographic Stratification.} Left: Red Knot. Right: Red Fox.}
    \label{fig:apdx:space_time}
\end{figure}

We also study a few more examples of geographic stratification in~\cref{fig:apdx:space_time}. 
First, the Red Knot has subspecies belonging to different geographical regions, similar to the Dunlin. However, the difference in bill proportions is not that significant here. Interestingly, we note that the sexual dimorphism is more pronounced fit the west-coast populations than their east coast counterparts~\cite{botwredknot}, and the bimodal nature of our estimates is also more pronounced for the west-coast population. Because we do not have ground truth genders for iNaturalist images, this is merely speculative, and a more detailed analysis is not currently feasible.

Next, we look at the Red Fox. Red foxes typically have larger ears in warmer climates for better body heat management. We see a trend where the ear : inter-eye proportion is higher for middle eastern populations than the populations in colder climates in Europe and North America. 
We also include a few examples of these species in~\cref{fig:apdx:pred_vis_spacetime}.

\subsection{Challenges in Measurement from Uncurated Images}
\label{sec:apdx:uncurated}

\begin{figure}[ht]
    \centering
    \includegraphics[width=\linewidth]{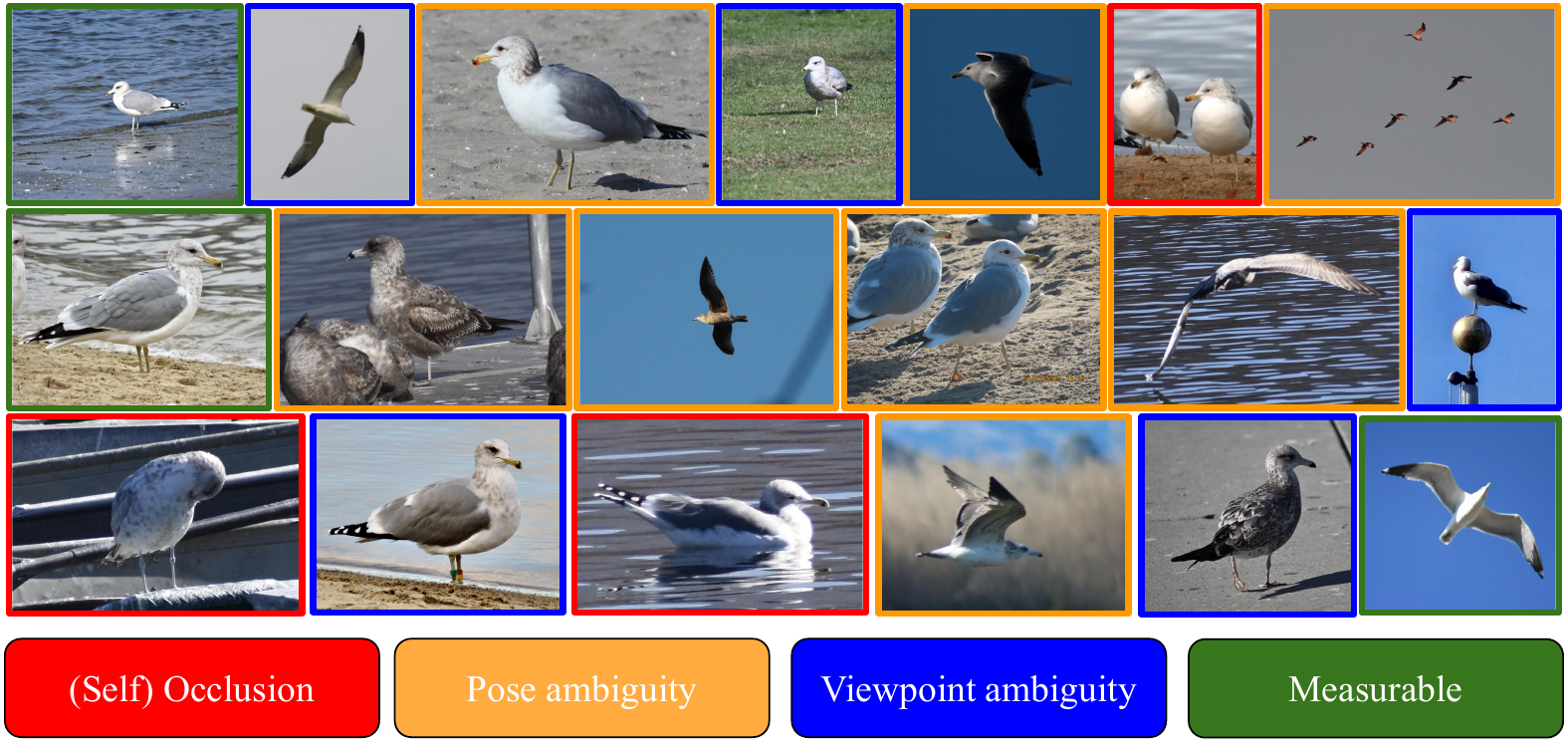}
    \caption{\textbf{Example images and measurability.} We show examples of California Gull images from iNaturalist, indicating whether each image is measurable and, when it is not, the primary limiting factor.}
    \label{fig:apdx:cal_gull}
\end{figure}

In~\cref{fig:apdx:cal_gull}, we show a number of randomly chosen images of California Gull from iNaturalist. Randomly chosen images from uncurated image collections may involve occlusion, pose ambiguity or viewpoint ambiguity. There are a fraction of images without these issues, however, which allow measurement. Our pose-aware retrieval strategy is intended to find such images. We present a number of images belonging to each category in \cref{fig:apdx:cal_gull}.

\begin{figure}
    \centering
    \includegraphics[width=\linewidth]{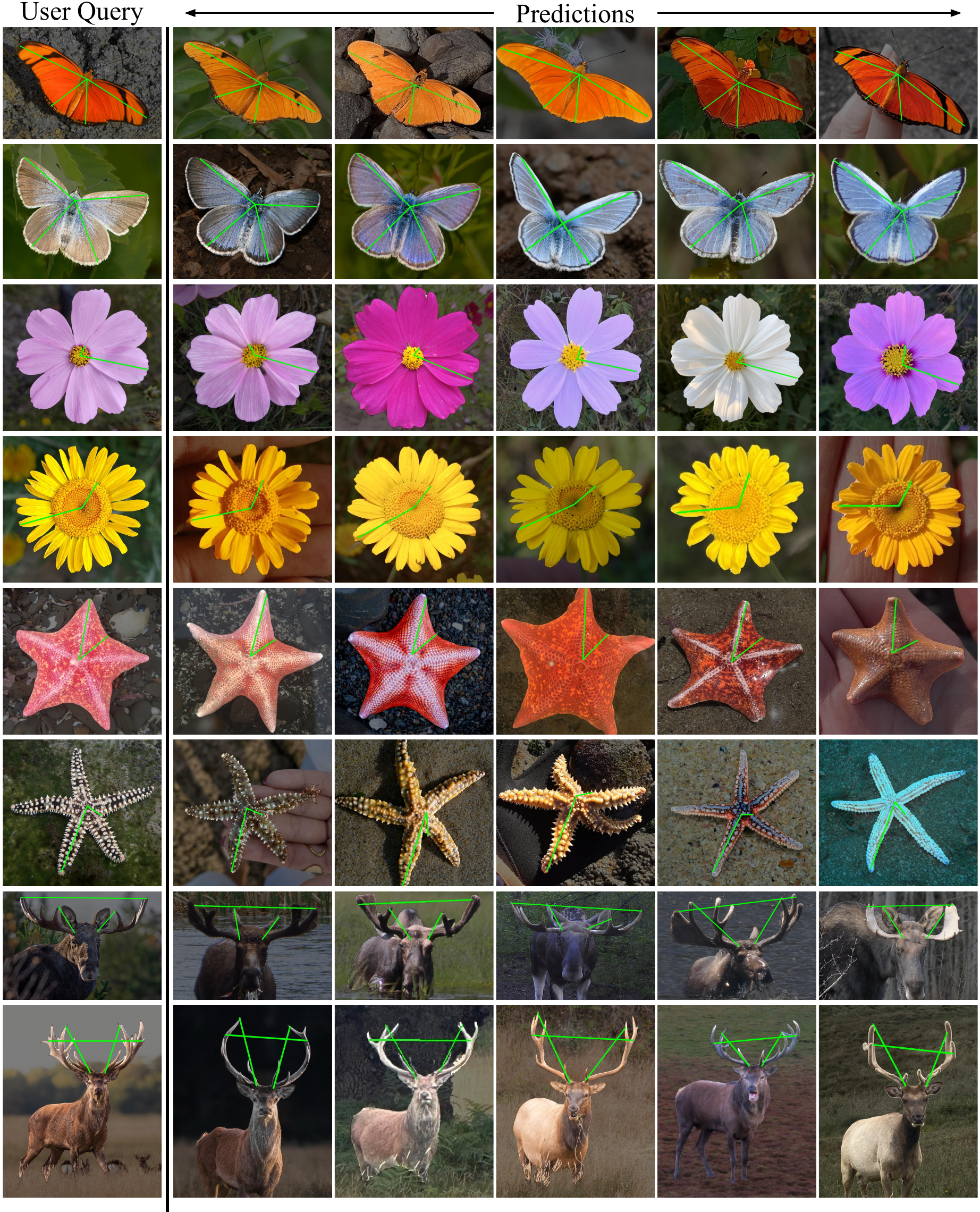}
    \caption{\textbf{Prediction visualizations for case studies.} Each row shows the user query with annotated keypoints, followed by five sample predictions. We show examples from the taxonomic groups used in our case studies, including butterflies, flowers, starfish, and deer.}
    \label{fig:apdx:pred_vis_cases}
\end{figure}

\begin{figure}
    \centering
    \includegraphics[width=\linewidth]{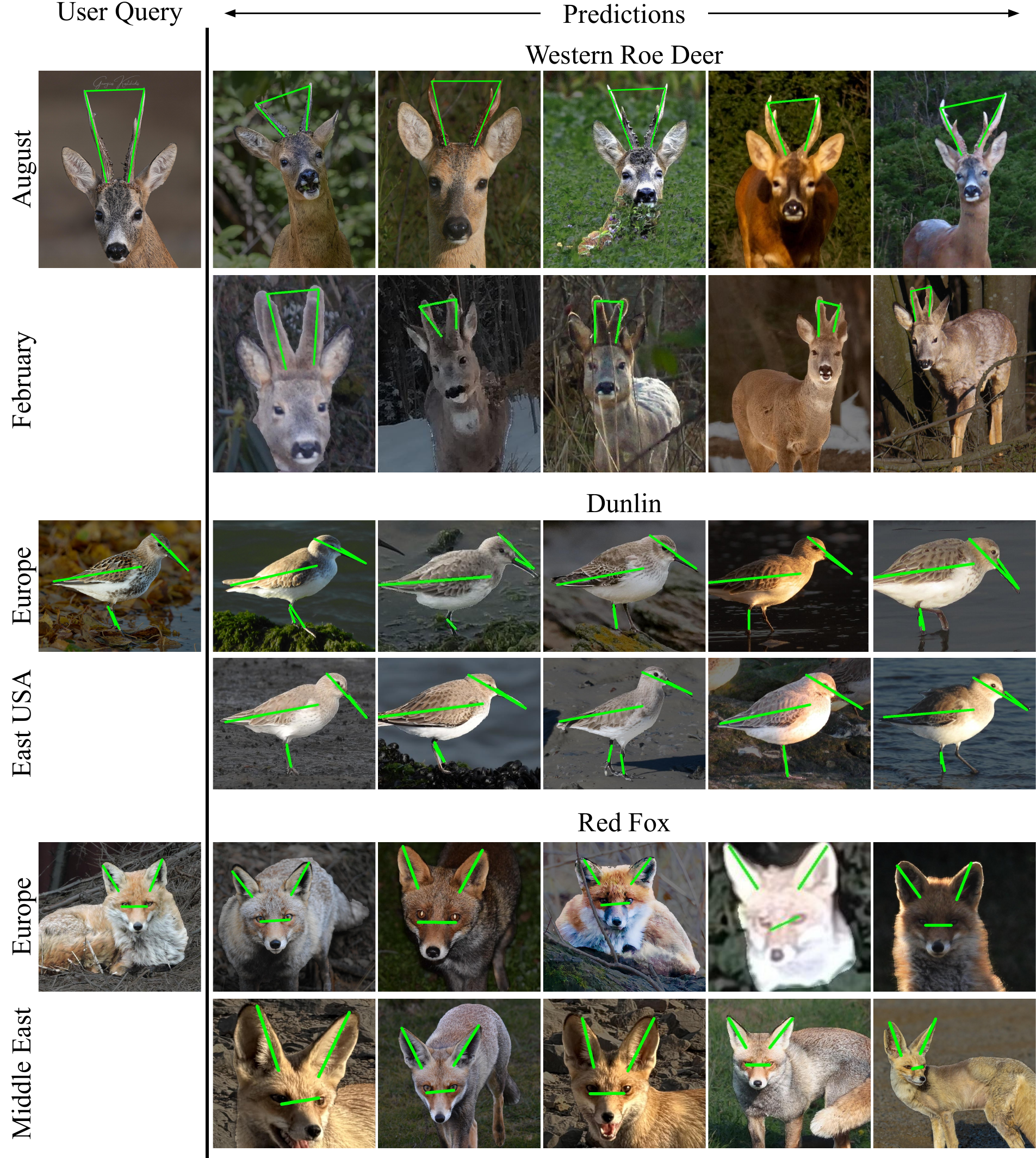}
    \caption{\textbf{Prediction visualizations for spatiotemporal stratification.} Each row shows the user query with annotated keypoints, followed by five sample predictions. We show examples from Western Roe Deer, Dunlin, and Red Fox, which are used in our stratification experiments. For each species, we compare two subgroups (\eg month or location) in separate rows. A single query image is used per species and is repeated next to each subgroup for ease of comparison.}
    \label{fig:apdx:pred_vis_spacetime}
\end{figure}

\subsection{Prediction Visualizations for Case Studies}
\label{sec:apdx:vis_case_studies}
We present selected predictions for species from our case study experiments in~\cref{fig:apdx:pred_vis_cases}. While broader trends across species within the same taxonomic group are generally apparent, there is also intra-species variation across individuals, particularly for butterfly wings and deer antlers. Garden cosmos, shown in the third row, also exhibits substantial visual variation, and the predictions demonstrate the robustness of \method to such variation.

Next, we present prediction visualizations for the spatiotemporal stratification experiments in~\cref{fig:apdx:pred_vis_spacetime}. Different subgroups are shown in separate rows to facilitate cross-subgroup comparison of estimated body proportions. February deer predictions generally show shorter antlers than August predictions, Dunlin bill differences are more subtle, and Middle Eastern red fox predictions tend to show slightly larger ears. These qualitative trends are consistent with the estimated proportion distributions presented in the main paper.

\end{document}